# Supervised Speech Separation Based on Deep Learning: An Overview

DeLiang Wang, *Fellow, IEEE,* and Jitong Chen

*Abstract*— Speech separation is the task of separating target speech from background interference. Traditionally, speech separation is studied as a signal processing problem. A more recent approach formulates speech separation as a supervised learning problem, where the discriminative patterns of speech, speakers, and background noise are learned from training data. Over the past decade, many supervised separation algorithms have been put forward. In particular, the recent introduction of deep learning to supervised speech separation has dramatically accelerated progress and boosted separation performance. This article provides a comprehensive overview of the research on deep learning based supervised speech separation in the last several years. We first introduce the background of speech separation and the formulation of supervised separation. Then we discuss three main components of supervised separation: learning machines, training targets, and acoustic features. Much of the overview is on separation algorithms where we review monaural methods, including speech enhancement (speech-nonspeech separation), speaker separation (multi-talker separation), and speech dereverberation, as well as multi-microphone techniques. The important issue of generalization, unique to supervised learning, is discussed. This overview provides a historical perspective on how advances are made. In addition, we discuss a number of conceptual issues, including what constitutes the target source.

## I. INTRODUCTION

The goal of speech separation is to separate target speech from background interference. Speech separation is a fundamental task in signal processing with a wide range of applications, including hearing prosthesis, mobile telecommunication, and robust automatic speech and speaker recognition. The human auditory system has the remarkable ability to extract one sound source from a mixture of multiple sources. In an acoustic environment like a cocktail party, we seem capable of effortlessly following one speaker in the presence of other speakers and background noises. Speech separation is commonly called the "cocktail party problem," a term coined by Cherry in his famous 1953 paper [26].

Speech separation is a special case of sound source separation. Perceptually, source separation corresponds to auditory stream segregation, a topic of extensive research in auditory perception. The first systematic study on stream segregation was conducted by Miller and Heise [124] who noted that listeners split a signal with two alternating sine-wave tones into two streams. Bregman and his colleagues have carried out a series of studies on the subject, and in a seminal book [15] he introduced the term auditory scene analysis (ASA) to refer to the perceptual process that segregates an acoustic mixture and groups the signal originating from the same sound source. Auditory scene analysis is divided into simultaneous organization and sequential organization. Simultaneous organization (or grouping) integrates concurrent sounds, while sequential organization integrates sounds across time. With auditory patterns displayed on a time-frequency representation such as a spectrogram, main organizational principles responsible for ASA include: Proximity in frequency and time, harmonicity, common amplitude and frequency modulation, onset and offset synchrony, common location, and prior knowledge (see among others [163] [15] [29] [11] [30] [32]). These grouping principles also govern speech segregation [201] [154] [31] [4] [49] [93]. From ASA studies, there seems to be a consensus that the human auditory system segregates and attends to a target sound, which can be a tone sequence, a melody, or a voice. More debatable is the role of auditory attention in stream segregation [17] [151] [148] [120]. In this overview, we use speech separation (or segregation) primarily to refer to the computational task of separating the target speech signal from a noisy mixture.

How well do we perform speech segregation? One way of quantifying speech perception performance in noise is to measure speech reception threshold, the required SNR level for a 50% intelligibility score. Miller [123] reviewed human intelligibility scores when interfered by a variety of tones, broadband noises, and other voices. Listeners were tested for their word intelligibility scores, and the results are shown in Figure 1. In general, tones are not as interfering as broadband

D. L. Wang is with the Department of Computer Science and Engineering and the Center for Cognitive and Brain Sciences, The Ohio State University, Columbus, OH 43210, USA (e-mail: dwang@cse.ohio-state.edu). He also holds a visiting appointment at the Center of Intelligent Acoustics and Immersive Communications, Northwestern Polytechnical University, Xi'an, China.

J. Chen was with the Department of Computer Science and Engineering, The Ohio State University, Columbus, OH 43210, USA (email: chen.2593@osu.edu). He is now with Silicon Valley AI Lab at Baidu Research, 1195 Bordeaux Drive, Sunnyvale, CA 94089, USA.

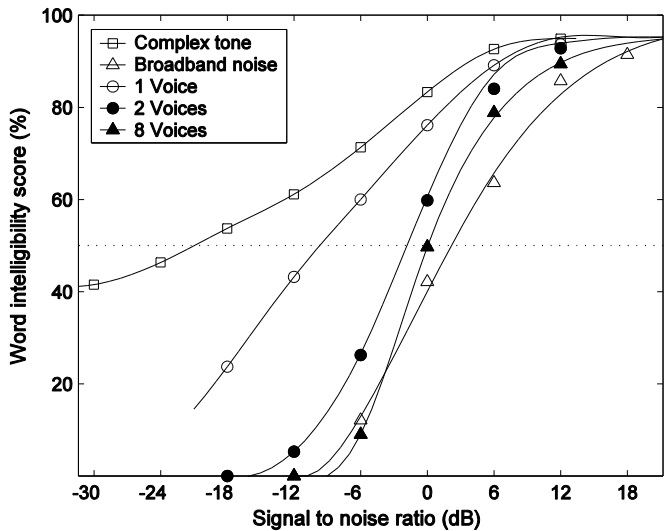

Figure 1. Word intelligibility score with respect to SNR for different kinds of interference (from [172], redrawn from [123]). The dashed line indicates 50% intelligibility. For speech interference, scores are shown for 1, 2, and 8 interfering speakers.

noises. For example, speech is intelligible even when mixed with a complex tone glide that is 20 dB more intense (pure tones are even weaker interferers). Broadband noise is the most interfering for speech perception, and the corresponding SRT is about 2 dB. When interference consists of other voices, the SRT depends on how many interfering talkers are present. As shown in Fig. 1, the SRT is about –10 dB for a single interferer but rapidly increases to –2 dB for two interferers. The SRT stays about the same (around –1 dB) when the interference contains four or more voices. There is a whopping SRT gap of 23 dB for different kinds of interference! Furthermore, it should be noted that listeners with hearing loss show substantially higher SRTs than normal-hearing listeners, ranging from a few decibels for broadband stationary noise to as high as 10-15 dB for interfering speech [44] [127], indicating a poorer ability of speech segregation.

With speech as the most important means of human communication, the ability to separate speech from background interference is crucial, as the speech of interest, or *target speech*, is usually corrupted by additive noises from other sound sources and reverberation from surface reflections. Although humans perform speech separation with apparent ease, it has proven to be very challenging to construct an automatic system to match the human auditory system in this basic task. In his 1957 book [27], Cherry made an observation: "No machine has yet been constructed to do just that [solving the cocktail part problem]." His conclusion, unfortunately for our field, has remained largely true for 6 more decades, although recent advances reviewed in this article have started to crack the problem.

Given the importance, speech separation has been extensively studied in signal processing for decades. Depending on the number of sensors or microphones, one can categorize separation methods into monaural (single-microphone) and array-based (multi-microphone). Two traditional approaches for monaural separation are speech enhancement [113] and computational auditory scene analysis (CASA) [172]. Speech enhancement analyzes general statistics of speech and noise, followed by estimation of clean speech from noisy speech with a noise estimate [40] [113]. The simplest and most widely used enhancement method is spectral subtraction [13], in which the power spectrum of the estimated noise is subtracted from that of noisy speech. In order to estimate background noise, speech enhancement techniques typically assume that background noise is stationary, i.e. its spectral properties do not change over time, or at least are more stationary than speech. CASA is based on perceptual principles of auditory scene analysis [15] and exploits grouping cues such as pitch and onset. For example, the tandem algorithm separates voiced speech by alternating pitch estimation and pitch-based grouping [78].

An array with two or more microphones uses a different principle to achieve speech separation. Beamforming, or spatial filtering, boosts the signal that arrives from a specific direction through proper array configuration, hence attenuating interference from other directions [164] [14] [9] [88]. The simplest beamformer is a delay-and-sum technique that adds multiple microphone signals from the target direction in phase and uses phase differences to attenuate signals from other directions. The amount of noise attenuation depends on the spacing, size, and configuration of the array – generally the attenuation increases as the number of microphones and the array length increase. Obviously, spatial filtering cannot be applied when target and interfering sources are co-located or near to one another. Moreover, the utility of beamforming is much reduced in reverberant conditions, which smear the directionality of sound sources.

A more recent approach treats speech separation as a supervised learning problem. The original formulation of supervised speech separation was inspired by the concept of time-frequency (T-F) masking in CASA. As a means of separation, T-F masking applies a two-dimensional mask (weighting) to the time-frequency representation of a source mixture in order to separate the target source [117] [172] [170]. A major goal of CASA is the ideal binary mask (IBM) [76], which denotes whether the target signal dominates a T-F unit in the time-frequency representation of a mixed signal. Listening studies show that ideal binary masking dramatically improves speech intelligibility for normal-hearing (NH) and hearing-impaired (HI) listeners in noisy conditions [16] [1] [109] [173]. With the IBM as the computational goal, speech separation becomes binary classification, an elementary form of supervised learning. In this case, the IBM is used as the desired signal, or target function, during training. During testing, the learning machine aims to estimate the IBM. Although it served as the first training target in supervised speech separation, the IBM is by no means the only training target and Sect. III presents a list of training targets, many shown to be more effective.

Since the formulation of speech separation as classification, the data-driven approach has been extensively studied in the speech processing community. Over the last decade, supervised speech separation has substantially advanced the state-of-the-art performance by leveraging large training data and increasing computing resources [21]. Supervised separation has especially benefited from the rapid rise in deep

learning – the topic of this overview. Supervised speech separation algorithms can be broadly divided into the following components: learning machines, training targets, and acoustic features. In this paper, we will first review the three components. We will then move to describe representative algorithms, where monaural and array-based algorithms will be covered in separate sections. As generalization is an issue unique to supervised speech separation, this issue will be treated in this overview.

Let us clarify a few related terms used in this overview to avoid potential confusion. We refer to *speech separation* or segregation as the general task of separating target speech from its background interference, which may include nonspeech noise, interfering speech, or both, as well as room reverberation. Furthermore, we equate speech separation and the cocktail party problem, which goes beyond the separation of two speech utterances originally experimented with by Cherry [26]. By *speech enhancement* (or *denoising*), we mean the separation of speech and nonspeech noise. If one is limited to the separation of multiple voices, we use the term *speaker separation*.

This overview is organized as follows. We first review the three main aspects of supervised speech separation, i.e., learning machines, training targets, and features, in Sections II, III, and IV, respectively. Section V is devoted to monaural separation algorithms, and Section VI to array-based algorithms. Section VII concludes the overview with a discussion of a few additional issues, such as what signal should be considered as the target and what a solution to the cocktail party problem may look like.

II. CLASSIFIERS AND LEARNING MACHINES

Over the past decade, DNNs have significantly elevated the performance of many supervised learning tasks, such as image classification [28], handwriting recognition [53], automatic speech recognition [73], language modeling [156], and machine translation [157]. DNNs have also advanced the performance of supervised speech separation by a large margin. This section briefly introduces the types of DNNs for supervised speech separation: feedforward multilayer perceptrons (MLPs), convolutional neural networks (CNNs), recurrent neural networks (RNNs), and generative adversarial networks (GANs).

The most popular model in neural networks is an MLP that has feedforward connections from the input layer to the output layer, layer-by-layer, and the consecutive layers are fully connected. An MLP is an extension of Rosenblatt's perceptron [142] by introducing hidden layers between the input layer and the output layer. An MLP is trained with the classical backpropagation algorithm [143] where the network weights are adjusted to minimize the prediction error through gradient descent. The prediction error is measured by a cost (loss) function between the predicted output and the desired output, the latter provided by the user as part of supervision. For example, when an MLP is used for classification, a popular cost function is cross entropy:

$$-\frac{1}{N}\sum_{i=1}^{N}\sum_{c=1}^{C} I_{i,c} log(p_{i,c})$$

where $i$ indexes an output model neuron and $p_{i,c}$ denotes the predicted probability of $i$ belonging to class $c$. $N$ and $C$ indicate the number of output neurons and the number of classes, respectively. $I_{i,c}$ is a binary indicator, which takes 1 if the desired class of neuron $i$ is $c$ and 0 otherwise. For function approximation or regression, a common cost function is mean square error (MSE):

$$\frac{1}{N}\sum_{i=1}^{N}(y_i - \hat{y}_i)^2$$

where $\hat{y}_i$ and $y_i$ are the predicted output and desired output for neuron $i$, respectively.

The representational power of an MLP increases as the number of layers increases [142] even though, in theory, an MLP with two hidden layers can approximate any function [70]. The backpropagation algorithm is applicable to an MLP of any depth. However, a deep neural network (DNN) with many hidden layers is difficult to train from a random initialization of connection weights and biases because of the so-called vanishing gradient problem, which refers to the observation that, at lower layers (near the input end), gradients calculated from backpropagated error signals from upper layers, become progressively smaller or vanishing. As a result of vanishing gradients, connection weights at lower layers are not modified much and therefore lower layers learn little during training. This explains why MLPs with a single hidden layer were the most widely used neural network prior to the advent of DNN.

A breakthrough in DNN training was made by Hinton et al. [74]. The key idea is to perform layerwise unsupervised pretraining with unlabeled data to properly initialize a DNN before supervised learning (or fine tuning) is performed with labeled data. More specifically, Hinton et al. [74] proposed restrictive Boltzmann machines (RBMs) to pretrain a DNN layer by layer, and RBM pretraining is found to improve subsequent supervised learning. A later remedy was to use a rectified linear unit (ReLU) [128] to replace the traditional sigmoid activation function, which converts a weighted sum of the inputs to a model neuron to the neuron's output. Recent practice shows that a moderately deep MLP with ReLUs can be effectively trained with large training data without unsupervised pretraining. Recently, skip connections have been introduced to facilitate the training of very deep MLPs [153] [62].

A class of feedforward networks, known as convolutional neural networks (CNNs) [106] [10], has been demonstrated to be well suited for pattern recognition, particularly in the visual domain. CNNs incorporate well-documented invariances in pattern recognition such as translation (shift) invariance. A typical CNN architecture is a cascade of pairs of a convolutional layer and a subsampling layer. A convolutional layer consists of multiple feature maps, each of which learns to extract a local feature regardless of its position in the previous layer through weight sharing: the neurons within the same module are constrained to have the same connection weights despite their different receptive fields. A receptive field of a neuron in this context denotes the local area of the previous layer that is connected to the neuron, whose operation of a weighted sum is akin to a convolution[1]. Each convolutional layer is followed by a subsampling layer that performs local averaging or maximization over the receptive fields of the neurons in the convolutional layer. Subsampling serves to reduce resolution and sensitivity to local variations. The use of weight sharing in CNN also has the benefit of cutting down the number of trainable parameters. Because a CNN incorporates domain knowledge in pattern recognition via its network structure, it can be better trained by the backpropagation algorithm despite the fact that a CNN is a deep network.

RNNs allow recurrent (feedback) connections, typically between hidden units. Unlike feedforward networks, which process each input sample independently, RNNs treat input samples as a sequence and model the changes over time. A speech signal exhibits strong temporal structure, and the signal within the current frame is influenced by the signals in the previous frames. Therefore, RNNs are a natural choice for learning the temporal dynamics of speech. We note that a RNN through its recurrent connections introduces the time dimension, which is flexible and infinitely extensible, a characteristic not shared by feedforward networks no matter how deep they are [169]; in a way, a RNN can be viewed a DNN with an infinite depth [146]. The recurrent connections are typically trained with backpropagation through time [187]. However, such RNN training is susceptible to the vanishing or exploding gradient problem [137]. To alleviate this problem, a RNN with long short-term memory (LSTM) introduces memory cells with gates to facilitate the information flow over time [75]. Specifically, a memory cell has three gates: input gate, forget gate and output gate. The forget gate controls how much previous information should be retained, and the input gate controls how much current information should be added to the memory cell. With these gating functions, LSTM allows relevant contextual information to be maintained in memory cells to improve RNN training.

Generative adversarial networks (GANs) were recently introduced with simultaneously trained models: a generative model *G* and a discriminative model *D* [52]. The generator *G* learns to model labeled data, e.g. the mapping from noisy speech samples to their clean counterparts, while the discriminator – usually a binary classifier – learns to discriminate between generated samples and target samples from training data. This framework is analogous to a two-player adversarial game, where minimax is a proven strategy [144]. During training, *G* aims to learn an accurate mapping so that the generated data can well imitate the real data so as to fool *D*; on the other hand, *D* learns to better tell the difference between the real data and synthetic data generated by *G*. Competition in this game, or adversarial learning, drives both models to improve their accuracy until generated samples are indistinguishable from real ones. The key idea of GANs is to use the discriminator to shape the loss function of the generator. GANs have recently been used in speech enhancement (see Sect. V.A).

In this overview, a DNN refers to any neural network with at least two hidden layers [10] [73], in contrast to popular learning machines with just one hidden layer such as commonly used MLPs, support vector machines (SVMs) with kernels, and Gaussian mixture models (GMMs). As DNNs get deeper in practice, with more than 100 hidden layers actually used, the depth required for a neural network to be considered a DNN can be a matter of a qualitative, rather than quantitative, distinction. Also, we use the term DNN to denote any neural network with a deep structure, whether it is feedforward or recurrent.

We should mention that DNN is not the only kind of learning machine that has been employed for speech separation. Alternative learning machines used for supervised speech separation include GMM [147] [97], SVM [55], and neural networks with just one hidden layer [91]. Such studies will not be further discussed in this overview as its theme is DNN based speech separation.

### III. TRAINING TARGETS

In supervised speech separation, defining a proper training target is important for learning and generalization. There are mainly two groups of training targets, i.e., masking-based targets and mapping-based targets. Masking-based targets describe the time-frequency relationships of clean speech to background interference, while mapping-based targets correspond to the spectral representations of clean speech. In this section, we survey a number of training targets proposed in the field.

Before reviewing training targets, let us first describe evaluation metrics commonly used in speech separation. A variety of metrics has been proposed in the literature, depending on the objectives of individual studies. These metrics can be divided into two classes: signal-level and perception-level. At the signal level, metrics aim to quantify the degrees of signal enhancement or interference reduction. In addition to the traditional SNR, speech distortion (loss) and noise residue in a separated signal can be individually measured [77] [113]. A prominent set of evaluation metrics comprises SDR (source-to-distortion ratio), SIR (source-to-interference ratio), and SAR (source-to-artifact ratio) [165].

As the output of a speech separation system is often consumed by the human listener, a lot of effort has been made to quantitatively predict how the listener perceives a separated signal. Because intelligibility and quality are two primary but different aspects of speech perception, objective metrics have been developed to separately evaluate speech intelligibility and speech quality. With the IBM's ability to elevate human speech intelligibility and its connection to the articulation index (AI) [114] – the classic model of speech perception – the HIT−FA rate has been suggested as an evaluation metric with the IBM as the reference [97]. HIT denotes the percent of speech-dominant T-F units in the IBM that is correctly classified and FA (false-alarm) refers to the percent of noise-dominant units that is incorrectly classified. The HIT−FA rate is found to be well correlated with speech intelligibility [97]. In recent years, the most commonly used intelligibility metric is STOI (short-time objective intelligibility), which measures the correlation between the short-time temporal envelopes of a reference (clean) utterance and a separated utterance [158] [89]. The value range of STOI is typically between 0 and 1, which can be interpreted as percent correct. Although STOI tends to overpredict intelligibility scores [64] [102], no alternative metric has been shown to consistently correlate with human intelligibility better. For speech quality, PESQ (perceptual evaluation of speech quality) is the standard metric [140] and recommended by the International Telecommunication Union (ITU) [87]. PESQ applies an auditory transform to produce a loudness spectrum, and compares the loudness spectra of a clean reference signal and a separated signal to produce a score in a range of -0.5 to 4.5, corresponding to the prediction of the perceptual MOS (mean opinion score).

---
[1] More straightforwardly a correlation.

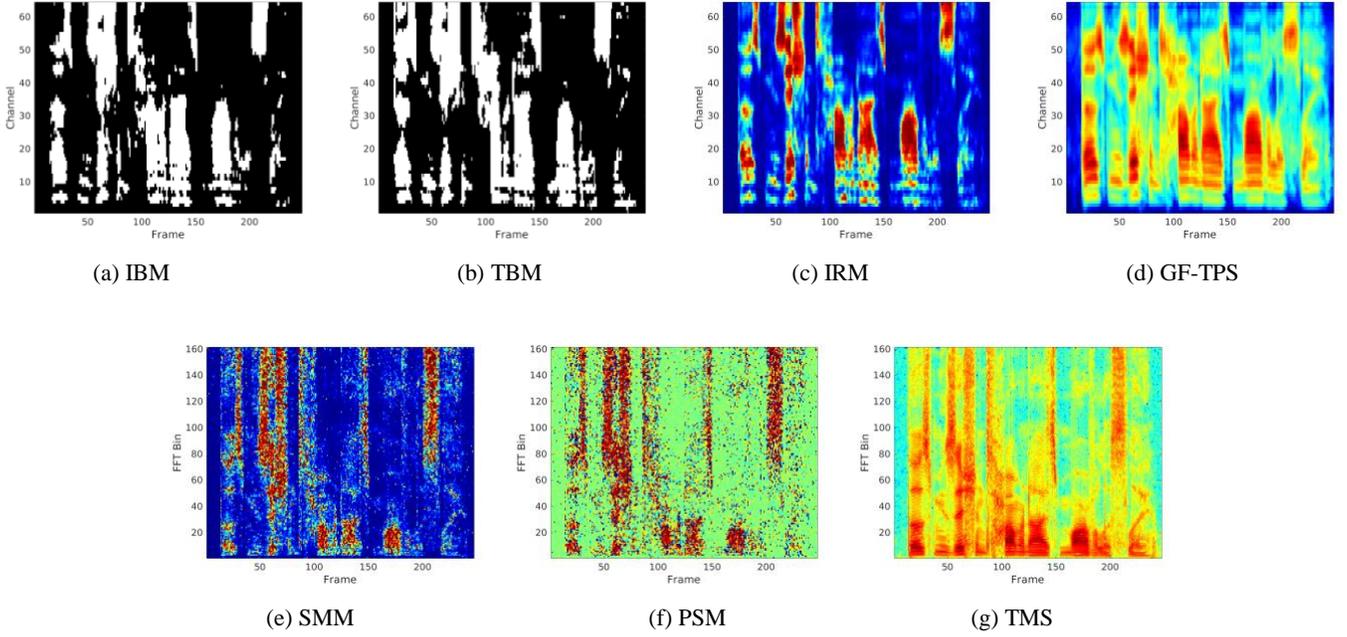

Figure 2. Illustration of various training targets for a TIMIT utterance mixed with a factory noise at -5 dB SNR.

### A. Ideal Binary Mask

The first training target used in supervised speech separation is the ideal binary mask [76] [141] [77] [168], which is inspired by the auditory masking phenomenon in audition [126] and the exclusive allocation principle in auditory scene analysis [15]. The IBM is defined on a two-dimensional T-F representation of a noisy signal, such as a cochleagram or a spectrogram:

$$IBM = \begin{cases} 1, & \text{if } SNR(t,f) > LC \\ 0, & \text{otherwise} \end{cases} \quad (1)$$

where $t$ and $f$ denote time and frequency, respectively. The IBM assigns the value 1 to a unit if the SNR within the T-F unit exceeds the local criterion (LC) or threshold, and 0 otherwise. Fig. 2(a) shows an example of the IBM, which is defined on a 64-channel cochleagram. As mentioned in Sect. I, IBM masking dramatically increases speech intelligibility in noise for normal-hearing and hearing-impaired listeners. The IBM labels every T-F unit as either target-dominant or interference-dominant. As a result, IBM estimation can naturally be treated as a supervised classification problem. A commonly used cost function for IBM estimation is cross entropy, as described in Section II.

### B. Target Binary Mask

Like the IBM, the target binary mask (TBM) categorizes all T-F units with a binary label. Different from the IBM, the TBM derives the label by comparing the target speech energy in each T-F unit with a fixed interference: speech-shaped noise, which is a stationary signal corresponding to the average of all speech signals. An example of the TBM is shown in Fig. 2(b). Target binary masking also leads to dramatic improvement of speech intelligibility in noise [99], and the TBM has been used as a training target [51] [112].

### C. Ideal Ratio Mask

Instead of a hard label on each T-F unit, the ideal ratio mask (IRM) can be viewed as a soft version of the IBM [152] [130] [178] [84]:

$$IRM = \left(\frac{S(t,f)^2}{S(t,f)^2 + N(t,f)^2}\right)^\beta \quad (2)$$

where $S(t,f)^2$ and $N(t,f)^2$ denote speech energy and noise energy within a T-F unit, respectively. The tunable parameter $\beta$ scales the mask, and is commonly chosen to 0.5. With the square root the IRM preserves the speech energy with each T-F unit, under the assumption that $S(t,f)$ and $N(t,f)$ are uncorrelated. This assumption holds well for additive noise, but not for convolutive interference as in the case of room reverberation (late reverberation, however, can be reasonably considered as uncorrelated interference.) Without the root the IRM in (2) is similar to the classical Wiener filter, which is the optimal estimator of target speech in the power spectrum. MSE is typically used as the cost function for IRM estimation. An example of the IRM is shown in Fig. 2(c).

### D. Spectral Magnitude Mask

The spectral magnitude mask (SMM) (called FFT-MASK in [178]) is defined on the STFT (short-time Fourier transform) magnitudes of clean speech and noisy speech:

$$SMM(t,f) = \frac{|S(t,f)|}{|Y(t,f)|} \quad (3)$$

where $|S(t,f)|$ and $|Y(t,f)|$ represent spectral magnitudes of clean speech and noisy speech, respectively. Unlike the IRM, the SMM is not upper-bounded by 1. To obtain separated speech, we apply the SMM or its estimate to the spectral magnitudes of noisy speech, and resynthesize separated speech with the phases of noisy speech (or an estimate of clean speech phases). Fig. 2(e) illustrates the SMM.

### E. Phase-Sensitive Mask

The phase-sensitive mask (PSM) extends the SMM by including a measure of phase [41]:

$$PSM(t,f) = \frac{|S(t,f)|}{|Y(t,f)|} \cos\theta \quad (4)$$

where $\theta$ denotes the difference of the clean speech phase and the noisy speech phase within the T-F unit. The inclusion of the phase difference in the PSM leads to a higher SNR, and tends to yield a better estimate of clean speech than the SMM [41]. An example of the PSM is shown in Fig. 2(f).

### F. Complex Ideal Ratio Mask

The complex ideal ratio mask (cIRM) is an ideal mask in the complex domain. Unlike the aforementioned masks, it can perfectly reconstruct clean speech from noisy speech [188]:

$$S = cIRM * Y \quad (5)$$

where $S, Y$ denote the STFT of clean speech and noisy speech, respectively, and '$*$' represents complex multiplication. Solving for mask components results in the following definition:

$$cIRM = \frac{Y_r S_r + Y_i S_i}{Y_r^2 + Y_i^2} + i\frac{Y_r S_i - Y_i S_r}{Y_r^2 + Y_i^2} \quad (6)$$

where $Y_r$ and $Y_i$ denote real and imaginary components of noisy speech, respectively, and $S_r$ and $S_i$ real and imaginary components of clean speech, respectively. The imaginary unit is denoted by '$i$'. Thus the cIRM has a real component and an imaginary component, which can be separately estimated in the real domain. Because of complex-domain calculations, mask values become unbounded. So some form of compression should be used to bound mask values, such as a tangent hyperbolic or sigmoidal function [188] [184].

Williamson et al. [188] observe that, in Cartesian coordinates, structure exists in both real and imaginary components of the cIRM, whereas in polar coordinates, structure exists in the magnitude spectrogram but not phase spectrogram. Without clear structure, direct phase estimation would be intractable through supervised learning, although we should mention a recent paper that uses complex-domain DNN to estimate complex STFT coefficients [107]. On the other hand, an estimate of the cIRM provides a phase estimate, a property not possessed by PSM estimation.

### G. Target Magnitude Spectrum

The target magnitude spectrum (TMS) of clean speech, or $|S(t,f)|$, is a mapping-based training target [116] [196] [57] [197]. In this case supervised learning aims to estimate the magnitude spectrogram of clean speech from that of noisy speech. Power spectrum, or other forms of spectra such as mel spectrum, may be used instead of magnitude spectrum, and a log operation is usually applied to compress the dynamic range and facilitate training. A prominent form of the TMS is the log-power spectrum normalized to zero mean and unit variance [197]. An estimated speech magnitude is then combined with noisy phase to produce the separated speech waveform. In terms of cost function, MSE is usually used for TMS estimation. Alternatively, maximum likelihood can be employed to train a TMS estimator that explicitly models output correlation [175]. Fig. 2(g) shows an example of the TMS.

### H. Gammatone Frequency Target Power Spectrum

Another closely related mapping-based target is the gammatone frequency target power spectrum (GF-TPS) [178]. Unlike the TMS defined on a spectrogram, this target is defined on a cochleagram based on a gammatone filterbank. Specifically, this target is defined as the power of the cochleagram response to clean speech. An estimate of the GF-TPS is easily converted to the separated speech waveform through cochleagram inversion [172]. Fig. 2(d) illustrates this target.

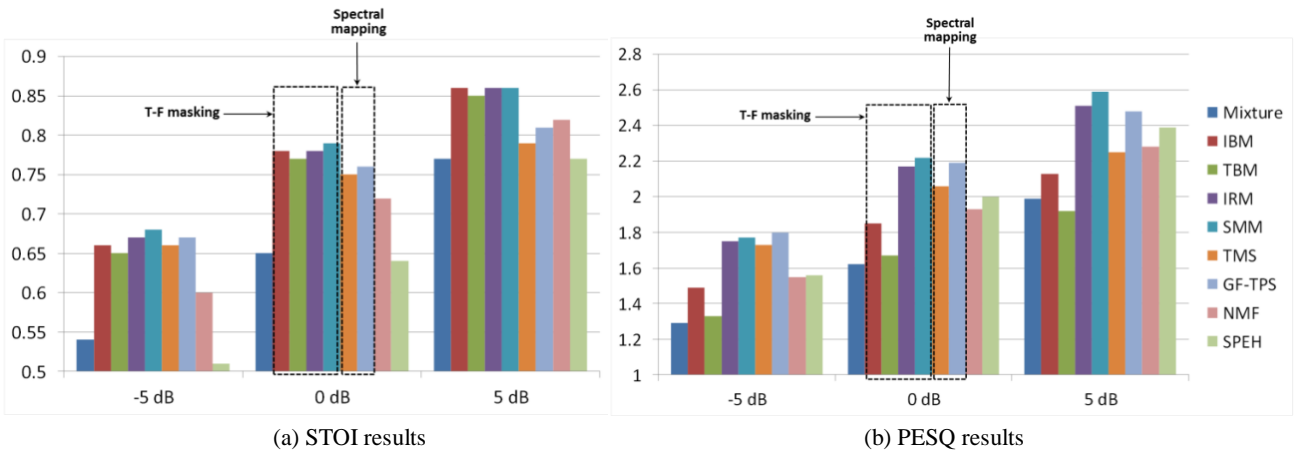

(a) STOI results  (b) PESQ results

Figure 3. Comparison of training targets. (a) In terms of STOI. (b) In terms of PESQ. Clean speech is mixed with a factory noise at -5 dB, 0 dB and 5 dB SNR. Results for different training targets as well as a speech enhancement (SPEH) algorithm and an NMF method are highlighted for 0 dB mixtures. Note that the results and the data in this figure can be obtained from a Matlab toolbox at http://web.cse.ohio-state.edu/pnl/DNN_toolbox/.

*I.  Signal Approximation*

The idea of signal approximation (SA) is to train a ratio mask estimator that minimizes the difference between the spectral magnitude of clean speech and that of estimated speech [186] [81]:

$$SA(t,f) = [RM(t,f)|Y(t,f)| - |S(t,f)|]^2 \qquad (7)$$

$RM(t,f)$ refers to an estimate of the SMM. So, SA can be interpreted as a target that combines ratio masking and spectral mapping, seeking to maximize SNR [186]. A related, earlier target aims for the maximal SNR in the context of IBM estimation [91]. For the SA target, better separation performance is achieved with two-stage training [186]. In the first stage, a learning machine is trained with the SMM as the target. In the second stage, the learning machine is fine-tuned by minimizing the loss function of (7).

A number of training targets have been compared using a fixed feedforward DNN with three hidden layers and the same set of input features [178]. The separated speech using various training targets is evaluated in terms of STOI and PESQ, for predicted speech intelligibility and speech quality, respectively. In addition, a representative speech enhancement algorithm [66] and a supervised nonnegative matrix factorization (NMF) algorithm [166] are evaluated as benchmarks. The evaluation results are given in Figure 3. A number of conclusions can be drawn from this study. First, in terms of objective intelligibility, the masking-based targets as a group outperform the mapping-based targets, although a recent study [155] indicates that masking is advantageous only at higher input SNRs and at lower SNRs mapping is more advantageous[2]. In terms of speech quality, ratio masking performs better than binary masking. Particularly illuminating is the contrast between the SMM and the TMS, which are the same except for the use of $|Y(t,f)|$ in the denominator of the SMM (see (3)). The better estimation of the SMM may be attributed to the fact that the target magnitude spectrum is insensitive to the interference signal and SNR, whereas the SMM is. The many-to-one mapping in the TMS makes its estimation potentially more difficult than SMM estimation. In addition, the estimation of unbounded spectral magnitudes tends to magnify estimation errors [178]. Overall, the IRM and the SMM emerge as the preferred targets. In addition, DNN based ratio masking performs substantially better than supervised NMF and unsupervised speech enhancement.

The above list of training targets is not meant to be exhaustive, and other targets have been used in the literature. Perhaps the most straightforward target is the waveform (time-domain) signal of clean speech. This indeed was used in an early study that trains an MLP to map from a frame of noisy speech waveform to a frame of clean speech waveform, which may be called *temporal mapping* [160]. Although simple, such direct mapping does not perform well even when a DNN is used in place of a shallow network [182] [34]. In [182], a target is defined in the time domain but the DNN for target estimation includes modules for ratio masking and inverse Fourier transform with noisy phase. This target is closely related to the PSM[3]. A recent study evaluates oracle results of a number of ideal masks and additionally introduces the so-called ideal gain mask (IGM) [184], defined in terms of *a priori* SNR and *a posteriori* SNR commonly used in traditional speech enhancement [113]. In [192], the so-called optimal ratio mask that takes into account of the correlation between target speech and background noise [110] was evaluated and found to be an effective target for DNN-based speech separation.

## IV. FEATURES

Features as input and learning machines play complementary roles in supervised learning. When features are discriminative, they place less demand on the learning machine in order to perform a task successfully. On the other hand, a powerful learning machine places less demand on features. At one extreme, a linear classifier, like Rosenblatt's perceptron, is all that is needed when features make a classification task linearly separable. At the other extreme, the input in the original form without any feature extraction (e.g. waveform in audio) suffices if the classifier is capable of learning appropriate features. In between are a majority of tasks where both feature extraction and learning are important.

Early studies in supervised speech separation use only a few features such as interaural time differences (ITD) and interaural level (intensity) differences (IID) [141] in binaural separation, and pitch-based features [91] [78] [55] and amplitude modulation spectrogram (AMS) [97] in monaural separation. A subsequent study [177] explores more monaural features including mel-frequency cepstral coefficient (MFCC), gammatone frequency cepstral coefficient (GFCC) [150], perceptual linear prediction (PLP) [67], and relative spectral transform PLP (RASTA-PLP) [68]. Through feature selection using group Lasso, the study recommends a complementary feature set comprising AMS, RASTA-PLP, and MFCC (and pitch if it can be reliably estimated), which has since been used in many studies.

We conducted a study to examine an extensive list of acoustic features for supervised speech separation at low SNRs [22]. The features have been previously used for robust automatic speech recognition and classification-based speech separation. The feature list includes mel-domain, linear-prediction, gammatone-domain, zero-crossing, autocorrelation, medium-time-filtering, modulation, and pitch-based features. The mel-domain features are MFCC and delta-spectral cepstral coefficient (DSCC) [104], which is similar to MFCC except that a delta operation is applied to mel-spectrum. The linear prediction features are PLP and RASTA-PLP. The three gammatone-domain features are gammatone feature (GF), GFCC, and gammatone frequency modulation coefficient (GFMC) [119]. GF is computed by passing an input signal to a gammatone filterbank and applying a decimation operation to subband signals. A zero-crossing feature, called zero-crossings with peak-amplitudes (ZCPA) [96], computes zero-crossing intervals and corresponding peak amplitudes from subband signals derived using a gammatone filterbank. The autocorrelation features are relative autocorrelation sequence MFCC (RAS-MFCC) [204], autocorrelation sequence MFCC (AC-MFCC) [149] and phase autocorrelation MFCC (PAC-MFCC) [86], all of which apply the MFCC procedure in the autocorrelation domain. The medium-time filtering features are power normalized cepstral coefficients (PNCC) [95] and suppression of slowly-varying components and the falling edge of the power envelope (SSF) [94]. The modulation domain features are Gabor filterbank (GFB) [145] and AMS features. Pitch-based (PITCH) features calculate T-F level features based on pitch tracking and use periodicity and instantaneous frequency to discriminate speech-dominant T-F units from noise-dominant ones. In addition to existing features, we proposed a new feature called Multi-Resolution Cochleagram (MRCG) [22], which computes four cochleagrams at different spectrotemporal resolutions to provide both local information and a broader context.

---

[2] The conclusion is also nuanced for speaker separation [206].

[3] This was first pointed out by Hakan Erdogan in personal communication.

The features are post-processed with the auto-regressive moving average (ARMA) filter [19] and evaluated with a fixed MLP based IBM mask estimator. The estimated masks are evaluated in terms of classification accuracy and the HIT−FA rate. The HIT−FA results are shown in Table 1. As shown in the table, gammatone-domain features (MRCG, GF, and GFCC) consistently outperform the other features in both accuracy and HIT−FA rate, with MRCG performing the best. Cepstral compaction via discrete cosine transform (DCT) is not effective, as revealed by comparing GF and GFCC features. Neither is modulation extraction, as shown by comparing GFCC and GMFC, the latter calculated from the former. It is worth noting that the poor performance of pitch features is largely due to inaccurate estimation at low SNRs, as ground-truth pitch is shown to be quite discriminative.

Recently, Delfarah and Wang [34] performed another feature study that considers room reverberation, and both speech denoising and speaker separation. Their study uses a fixed DNN trained to estimate the IRM, and the evaluation results are given in terms of STOI improvements over unprocessed noisy and reverberant speech. The features added in this study include log spectral magnitude (LOG-MAG) and log mel-spectrum feature (LOG-MEL), both of which are commonly used in supervised separation [196] [82]. Also included is waveform signal (WAV) without any feature extraction. For reverberation, simulated room impulse responses (RIRs) and recorded RIRs are both used with reverberation time up to 0.9 seconds. For denoising, evaluation is done separately for matched noises where the first half of each nonstationary noise is used in training and second half for testing, and unmatched noises where completely new noises are used for testing. For cochannel (two-speaker) separation, the target talker is male while the interfering talker is either female or male. Table 2 shows the STOI gains for the individual features evaluated. In the anechoic, matched noise case, STOI results are largely consistent with Table 1. Feature results are also broadly consistent using simulated and recorded RIRs. However, the best performing features are different for the matched noise, unmatched noise, and speaker separation cases. Besides MRCG, PNCC and GFCC produce the best results for the unmatched noise and cochannel condition, respectively. For feature combination, this study concludes that the most effective feature set consists of PNCC, GF, and LOG-MEL for speech enhancement, and PNCC, GFCC, and LOG-MEL for speaker separation.

The large performance differences caused by features in both Table 1 and Table 2 demonstrate the importance of features for supervised speech separation. The inclusion of raw waveform signal in Table 2 further suggests that, without feature extraction, separation results are poor. But it should be noted that, the feedforward DNN used in [34] may not couple well with waveform signals, and CNNs and RNNs may be better suited for so-called end-to-end separation. We will come to this issue later.

Table 1. Classification performance of a list of acoustic features in terms of HIT−FA (in %) for six noises at -5 dB SNR, where FA is shown in parentheses (from [22]). Boldtype indicates best scores.

|  | Factory | Babble | Engine | Cockpit | Vehicle | Tank | Average |
|---|---|---|---|---|---|---|---|
| MRCG | **63** (7) | **49** (13) | **77** (4) | **73** (4) | **80** (10) | **77** (6) | **70** (7) |
| GF | 61 (7) | 45 (15) | 75 (4) | 71 (3) | 80 (10) | 76 (6) | 68 (8) |
| GFCC | 61 (6) | 46 (14) | 73 (4) | 70 (3) | 78 (11) | 74 (6) | 67 (7) |
| DSCC | 56 (7) | 42 (14) | 70 (5) | 66 (3) | 77 (11) | 73 (6) | 64 (8) |
| MFCC | 57 (7) | 43 (14) | 69 (5) | 67 (4) | 77 (11) | 72 (7) | 64 (8) |
| PNCC | 56 (6) | 44 (14) | 69 (5) | 66 (4) | 77 (11) | 71 (7) | 64 (8) |
| PLP | 56 (6) | 41 (12) | 68 (5) | 66 (4) | 77 (11) | 71 (7) | 63 (8) |
| AC-MFCC | 56 (6) | 42 (14) | 67 (5) | 65 (4) | 77 (11) | 71 (7) | 63 (8) |
| RAS-MFCC | 57 (6) | 41 (14) | 68 (5) | 66 (4) | 76 (11) | 71 (7) | 63 (8) |
| GFB | 57 (7) | 41 (18) | 67 (5) | 66 (4) | 75 (12) | 70 (7) | 63 (9) |
| ZCPA | 55 (8) | 40 (16) | 68 (5) | 65 (4) | 75 (13) | 70 (8) | 62 (9) |
| SSF | 54 (7) | 39 (15) | 67 (5) | 60 (4) | 76 (11) | 69 (7) | 61 (8) |
| RASTA-PLP | 52 (6) | 38 (15) | 64 (5) | 61 (4) | 76 (12) | 67 (7) | 60 (8) |
| GFMC | 48 (7) | 35 (15) | 61 (6) | 60 (5) | 67 (17) | 59 (9) | 55 (10) |
| PITCH | 46 (3) | 29 (22) | 50 (5) | 50 (2) | 59 (16) | 53 (7) | 48 (9) |
| AMS | 40 (6) | 27 (9) | 49 (5) | 52 (4) | 50 (31) | 45 (11) | 44 (11) |
| PAC-MFCC | 17 (5) | 11 (8) | 30 (9) | 29 (7) | 40 (48) | 21 (17) | 25 (16) |

Table 2. STOI improvements (in %) for a list of features averaged on a set of test noises (from [34]). "Sim." and "Rec." indicate simulated and recorded room impulse responses. Boldface indicates the best scores in each condition. In cochannel (two-talker) cases, the performance is shown separately for a female interferer and male interferer (in parentheses) with a male target talker.

| Feature | Matched noise | | | Unmatched noise | | | Cochannel | | | Average |
|---|---|---|---|---|---|---|---|---|---|---|
| | Anechoic | Sim. RIRs | Rec. RIRs | Anechoic | Sim. RIRs | Rec. RIRs | Anechoic | Sim. RIRs | Rec. RIRs | |
| MRCG | **7.12** | **14.25** | **12.15** | **7.00** | 7.28 | 8.99 | 21.25(13.00) | 22.93 (13.19) | 21.29 (12.81) | **12.92** |
| GF | 6.19 | 13.10 | 11.37 | 6.71 | 7.87 | 8.24 | 22.56(11.87) | 23.95 (12.31) | 22.35 (12.87) | 12.71 |
| GFCC | 5.33 | 12.56 | 10.99 | 6.32 | 6.92 | 7.01 | **23.53 (14.34)** | **23.95 (14.01)** | **22.76 (13.90)** | 12.50 |
| LOG-MEL | 5.14 | 12.07 | 10.28 | 6.00 | 6.98 | 7.52 | 21.18 (13.88) | 22.75 (13.54) | 21.71 (13.18) | 12.08 |
| LOG-MAG | 4.86 | 12.13 | 9.69 | 5.75 | 6.64 | 7.19 | 20.82 (13.84) | 22.57 (13.40) | 21.82 (13.55) | 11.91 |
| GFB | 4.99 | 12.47 | 11.51 | 6.22 | 7.01 | 7.86 | 19.61 (13.34) | 20.86 (11.97) | 19.97 (11.60) | 11.75 |
| PNCC | 1.74 | 8.88 | 10.76 | 2.18 | **8.68** | **10.52** | 19.97 (10.73) | 19.47 (10.03) | 19.35 (9.56) | 10.78 |
| MFCC | 4.49 | 11.03 | 9.69 | 5.36 | 5.96 | 6.26 | 19.82 (11.98) | 20.32 (11.47) | 19.66 (11.54) | 10.72 |
| RAS-MFCC | 2.61 | 10.47 | 9.56 | 3.08 | 6.74 | 7.37 | 18.12 (11.38) | 19.07 (11.19) | 17.87 (10.30) | 10.44 |
| AC-MFCC | 2.89 | 9.63 | 8.89 | 3.31 | 5.61 | 5.91 | 18.66 (12.50) | 18.64 (11.59) | 17.73 (11.27) | 9.87 |
| PLP | 3.71 | 10.36 | 9.10 | 4.39 | 5.03 | 5.81 | 16.84 (11.29) | 16.73 (10.92) | 15.46 (9.50) | 9.46 |
| SSF-II | 3.41 | 8.57 | 8.68 | 4.18 | 5.45 | 6.00 | 16.76 (10.07) | 17.72 (9.18) | 18.07 (8.93) | 9.09 |
| SSF-I | 3.31 | 8.35 | 8.53 | 4.09 | 5.17 | 5.77 | 16.25 (10.44) | 17.70 (9.40) | 18.04 (9.35) | 8.97 |
| RASTA-PLP | 1.79 | 7.27 | 8.56 | 1.97 | 6.62 | 7.92 | 11.03 (5.74) | 10.96 (6.06) | 10.27 (6.28) | 7.46 |
| PITCH | 2.35 | 4.62 | 4.79 | 3.36 | 3.36 | 4.61 | 19.71 (9.37) | 17.82 (8.45) | 16.87 (6.72) | 7.03 |
| GFMC | -0.68 | 7.05 | 5.00 | -0.54 | 4.44 | 4.16 | 5.04 (-0.07) | 6.01 (0.33) | 4.97 (0.28) | 4.40 |
| WAV | 0.94 | 2.32 | 2.68 | 0.02 | 0.99 | 1.63 | 11.62 (4.81) | 11.92 (6.25) | 10.54 (1.05) | 3.89 |
| AMS | 0.31 | 0.30 | -1.38 | 0.19 | -2.99 | -3.40 | 11.73 (5.96) | 10.97 (6.76) | 10.20 (4.90) | 1.71 |
| PAC-MFCC | 0.00 | -0.33 | -0.82 | 0.18 | -0.92 | -0.67 | 0.95 (0.15) | 1.25 (0.26) | 1.17 (0.09) | -0.17 |

## V. MONAURAL SEPARATION ALGORITHMS

In this section, we discuss monaural algorithms for speech enhancement, speech dereverberation as well as dereverberation plus denoising, and speaker separation. We explain representative algorithms and discuss generalization of supervised speech separation.

### A. Speech Enhancement

To our knowledge, deep learning was first introduced to speech separation by Wang and Wang in 2012 in two conference papers [179] [180], which were later extended to a journal version in 2013 [181]. They used DNN for subband classification to estimate the IBM. In the conference versions, feedforward DNNs with RBM pretraining were used as binary classifiers, as well as feature encoders for structured perceptrons [179] and conditional random fields [180]. They reported strong separation results in all cases of DNN usage, with better results for DNN used for feature learning due to the incorporation of temporal dynamics in structured prediction.

In the journal version [181], the input signal is passed through a 64-channel gammatone filterbank to derive subband signals, from which acoustic features are extracted within each T-F unit. These features form the input to subband DNNs (64 in total) to learn more discriminative features. This use of DNN for speech separation is illustrated in Figure 4. After DNN training, input features and learned features of the last hidden layer are concatenated and fed to linear SVMs to estimate the subband IBM efficiently. This algorithm was further extended to a two-stage DNN [65], where the first stage is trained to estimate the subband IBM as usual and the second stage explicitly incorporates the T-F context in the following way. After the first-stage DNN is trained, a unit-level output before binarization can be interpreted as the posterior probability that speech dominates the T-F unit. Hence the first-stage DNN output is considered a posterior mask. In the second stage, a T-F unit takes as input a local window of the posterior mask centered at the unit. The two-stage DNN is illustrated in Fig. 5. This second-stage structure is reminiscent of a convolutional layer in CNN but without weight sharing. This way of leveraging contextual information is shown to significantly improve classification accuracy. Subject tests demonstrate that this DNN produced large intelligibility improvements for both HI and NH listeners, with HI listeners benefiting more [65]. This is the first monaural algorithm to provide substantial speech intelligibility improvements for HI listeners in background noise, so much so that HI subjects with separation outperformed NH subjects without separation.

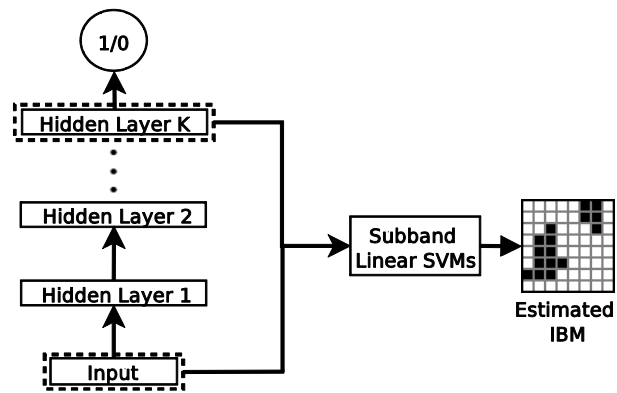

Figure 4. Illustration of DNN for feature learning, and learned features are then used by linear SVM for IBM estimation (from [181]).

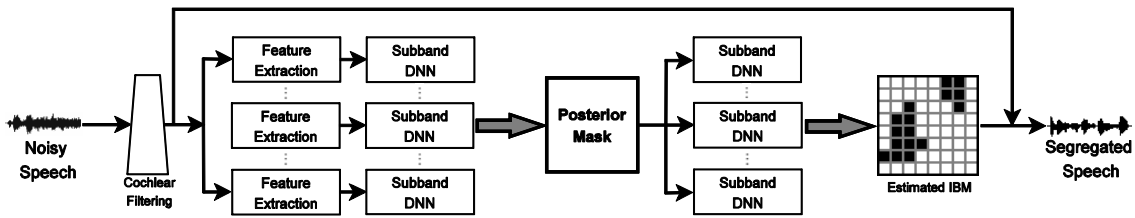

Figure 5. Schematic diagram of a two-stage DNN for speech separation (from [65]).

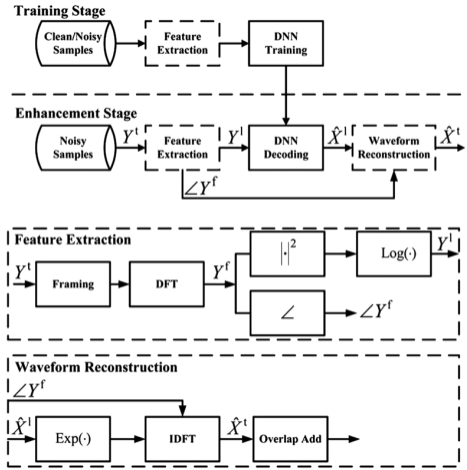

Figure 6. Diagram of a DNN-based spectral mapping method for speech enhancement (from [196]). The feature extraction and waveform reconstruction modules are further detailed.

In 2013, Lu et al. [116] published an Interspeech paper that uses a deep autoencoder (DAE) for speech enhancement. A basic autoencoder (AE) is an unsupervised learning machine, typically having a symmetric architecture with one hidden layer with tied weights, that learns to map an input signal to itself. Multiple trained AEs can be stacked into a DAE that is then subject to traditional supervised fine-tuning, e.g. with a backpropagation algorithm. In other words, autoencoding is an alternative to RBM pretraining. The algorithm in [116] learns to map from the mel-frequency power spectrum of noisy speech to that of clean speech, so it can be regarded as the first mapping based method[4].

Subsequently, but independent of [116], Xu et al. [196] published a study using a DNN with RBM pretraining to map from the log power spectrum of noisy speech to that of clean speech, as shown in Fig. 6. Unlike [116], the DNN used in [196] is a standard feedforward MLP with RBM pretraining. After training, DNN estimates clean speech's spectrum from a noisy input. Their experimental results show that the trained DNN yields about 0.4 to 0.5 PESQ gains over noisy speech on untrained noises, which are higher than those obtained by a representative traditional enhancement method.

Many subsequent studies have since been published along the lines of T-F masking and spectral mapping. In [186] [185], RNNs with LSTM were used for speech enhancement and its application to robust ASR, where training aims for signal approximation (see Sect. III.I). RNNs were also used in [41] to estimate the PSM. In [132] [210], a deep stacking network was proposed for IBM estimation and a mask estimate was then used for pitch estimation. The accuracy of both mask estimation and pitch estimation improves after the two modules iterate for several cycles. A DNN was used to simultaneously estimate the real and imaginary components of the cIRM, yielding better speech quality over IRM estimation [188]. Speech enhancement at the phoneme level has been recently studied [183] [18]. In [59], the DNN takes into account of perceptual masking with a piecewise gain function. In [198], multi-objective learning is shown to improve enhancement performance. It has been demonstrated that a hierarchical DNN performing subband spectral mapping yields better enhancement than a single DNN performing fullband mapping [39]. In [161], skip connections between non-consecutive layers are added to DNN to improve enhancement performance. Multi-target training with both masking and mapping based targets is found to outperform single-target training [205]. CNNs have also been used for IRM estimation [83] and spectral mapping [46] [136, 138].

---

[4] The authors also published a paper in Interspeech 2012 [115] where a DAE is trained in an unsupervised fashion to map from the mel-spectrum of clean speech to itself. The trained DAE is then used to "recall" a clean signal from a noisy input for robust ASR.

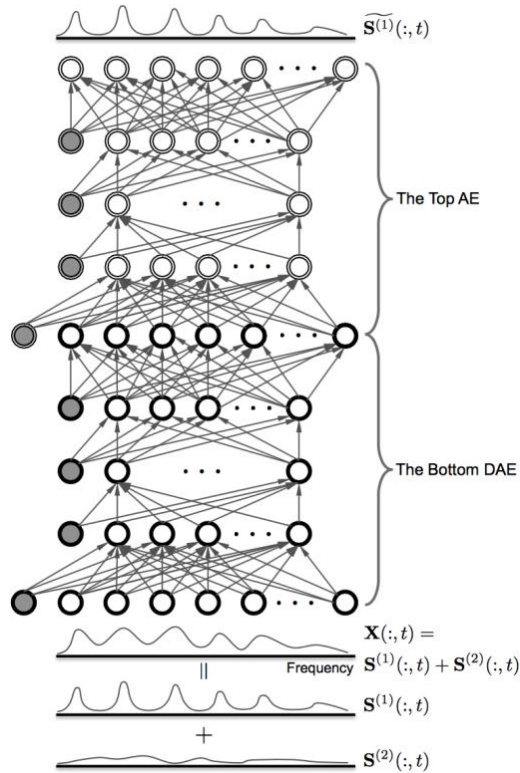

Figure 7. DNN architecture for speech enhancement with an autoencoder for unsupervised adaptation (from [98]). The AE stacked on top of a DNN serves as a purity checker for estimated clean speech from the bottom DNN. $S^{(1)}$ denotes the spectrum of a speech signal, $S^{(2)}$ the spectrum of a noise signal, and $\widetilde{S^{(1)}}$ an estimate of $S^{(1)}$.

Aside from masking and mapping based approaches, there is recent interest in using deep learning to perform end-to-end separation, i.e. temporal mapping without resorting to a T-F representation. A potential advantage of this approach is to circumvent the need to use the phase of noisy speech in reconstructing enhanced speech, which can be a drag for speech quality, particularly when input SNR is low. Recently, Fu et al. [47] developed a fully convolutional network (a CNN with fully connected layers removed) for speech enhancement. They observe that full connections make it difficult to map both high and low frequency components of a waveform signal, and with their removal, enhancement results improve. As a convolution operator is the same as a filter or a feature extractor, CNNs appear to be a natural choice for temporal mapping.

A recent study employs a GAN to perform temporal mapping [138]. In the so-called speech enhancement GAN (SEGAN), the generator is a fully convolutional network, performing enhancement or denoising. The discriminator follows the same convolutional structure as $G$, and it transmits information of generated waveform signals versus clean signals back to $G$. $D$ can be viewed as providing a trainable loss function for $G$. SEGAN was evaluated on untrained noisy conditions, but the results are inconclusive and worse than masking or mapping methods. In another GAN study [122], $G$ tries to enhance the spectrogram of noisy speech while $D$ tries to distinguish between the enhanced spectrograms and those of clean speech. The comparisons in [122] show that the enhancement results by this GAN are comparable to those achieved by a DNN.

Not all deep learning based speech enhancement methods build on DNNs. For example, Le Roux et al. [105] proposed deep NMF that unfolds NMF operations and includes multiplicative updates in backpropagation. Vu et al. [167] presented an NMF framework in which a DNN is trained to map NMF activation coefficients of noisy speech to their clean version.

### B. Generalization of Speech Enhancement Algorithms

For any supervised learning task, generalization to untrained conditions is a crucial issue. In the case of speech enhancement, data-driven algorithms bear the burden of proof when it comes to generalization, because the issue does not arise in traditional speech enhancement and CASA algorithms which make minimal use of supervised training. Supervised enhancement has three aspects of generalization: noise, speaker, and SNR. Regarding SNR generalization, one can simply include more SNR levels in a training set and practical experience shows that supervised enhancement is not sensitive to precise SNRs used in training. Part of the reason is that, even though a few mixture SNRs are included in training, local SNRs at the frame level and T-F unit level usually vary over a wide range, providing a necessary variety for a learning machine to generalize well. An alternative strategy is to adopt progressive training with increasing numbers of hidden layers

Table 3. Speech enhancement results at -2 dB SNR measured in STOI (from [24]).

|  | Babble1 | Cafeteria | Factory | Babble2 | Average |
| --- | --- | --- | --- | --- | --- |
| Unprocessed | 0.612 | 0.596 | 0.611 | 0.611 | 0.608 |
| 100-noise model | 0.683 | 0.704 | 0.750 | 0.688 | 0.706 |
| 10K-noise model | 0.792 | 0.783 | 0.807 | 0.786 | 0.792 |
| Noise-dependent model | 0.833 | 0.770 | 0.802 | 0.762 | 0.792 |

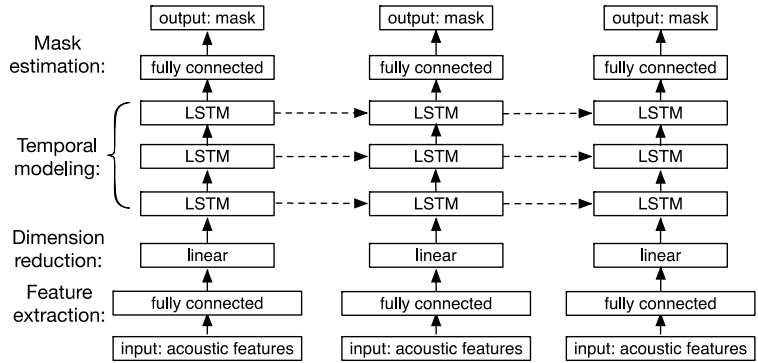

Figure 8. Diagram of an LSTM based speech separation system (from [20]).

to handle lower SNR conditions [48].

In an effort to address the mismatch between training and test conditions, Kim and Smaragdis [98] proposed a two-stage DNN where the first stage is a standard DNN to perform spectral mapping and the second stage is an autoencoder that performs unsupervised adaptation during the test stage. The AE is trained to map the magnitude spectrum of a clean utterance to itself, much like [115], and hence its training does not need labeled data. The AE is then stacked on top of the DNN, and serves as a purity checker as shown in Fig. 7. The rationale is that well enhanced speech tends to produce a small difference (error) between the input and the output of the AE, whereas poorly enhanced speech should produce a large error. Given a test mixture, the already-trained DNN is fine-tuned with the error signal coming from the AE. The introduction of an AE module provides a way of unsupervised adaptation to test conditions that are quite different from the training conditions, and is shown to improve the performance of speech enhancement.

Noise generalization is fundamentally challenging as all kinds of stationary and nonstationary noises may interfere with a speech signal. When available training noises are limited, one technique is to expand training noises through noise perturbation, particularly frequency perturbation [23]; specifically, the spectrogram of an original noise sample is perturbed to generate new noise samples. To make the DNN-based mapping algorithm of Xu et al. [196] more robust to new noises, Xu et al. [195] incorporate noise aware training, i.e. the input feature vector includes an explicit noise estimate. With noise estimated via binary masking, the DNN with noise aware training generalizes better to untrained noises.

Noise generalization is systematically addressed in [24]. The DNN in this study was trained to estimate the IRM at the frame level. In addition, the IRM is simultaneously estimated over several consecutive frames and different estimates for the same frame are averaged to produce a smoother, more accurate mask (see also [178]). The DNN has five hidden layers with 2048 ReLUs in each. The input features for each frame are cochleagram response energies (the GF feature in Tables 1 and 2). The training set includes 640,000 mixtures created from 560 IEEE sentences and 10,000 (10K) noises from a sound effect library (www.sound-ideas.com) at the fixed SNR of -2 dB. The total duration of the noises is about 125 hours, and the total duration of training mixtures is about 380 hours. To evaluate the impact of the number of training noises on noise generalization, the same DNN is also trained with 100 noises as done in [181]. The test sets are created using 160 IEEE sentences and nonstationary noises at various SNRs. Neither test sentences nor test noises are used during training. The separation results measured in STOI are shown in Table 3, and large STOI improvements are obtained by the 10K-noise model. In addition, the 10K-noise model substantially outperforms the 100-noise model, and its average performance matches the noise-dependent models trained with the first half of the training noises and tested with the second half. Subject tests show that the noise-independent model resulting from large-scale training significantly improves speech intelligibility for NH and HI listeners in unseen noises. This study strongly suggests that large-scale training with a wide variety of noises is a promising way to address noise

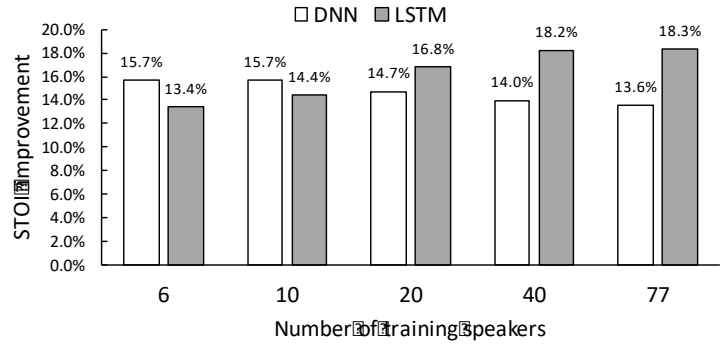

(a) Results for trained speakers at -5 dB SNR.

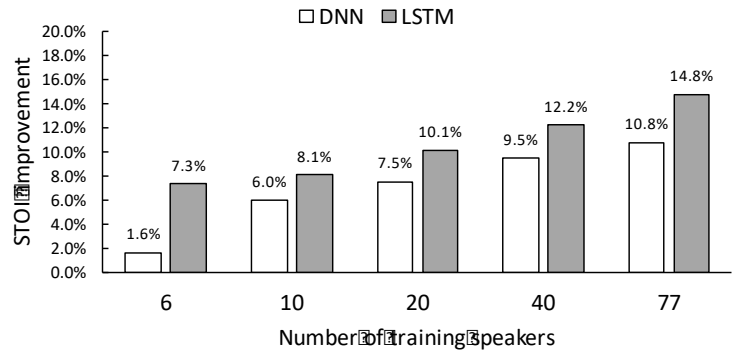

(b) Results for untrained speakers at -5 dB SNR.

Figure 9. STOI improvements of a feedforward DNN and a RNN with LSTM (from [20]).

generalization.

As for speaker generalization, a separation system trained on a specific speaker would not work well for a different speaker. A straightforward attempt for speaker generalization would be to train with a large number of speakers. However, experimental results [20] [100] show that a feedforward DNN appears incapable of modeling a large number of talkers. Such a DNN typically takes a window of acoustic features for mask estimation, without using the long-term context. Unable to track a target speaker, a feedforward network has a tendency to mistake noise fragments for target speech. RNNs naturally model temporal dependencies, and are thus expected to be more suitable for speaker generalization than feedforward DNN.

We have recently employed RNN with LSTM to address speaker generalization of noise-independent models [20]. The model, shown in Figure 8, is trained on 3,200,000 mixtures created from 10,000 noises mixed with 6, 10, 20, 40, and 77 speakers. When tested with trained speakers, as shown in Fig. 9(a), the performance of the DNN degrades as more training speakers are added to the training set, whereas the LSTM benefits from additional training speakers. For untrained test speakers, as shown in Fig. 9(b), the LSTM substantially outperforms the DNN in terms of STOI. LSTM appears able to track a target speaker over time after being exposed to many speakers during training. With large-scale training with many speakers and numerous noises, RNNs with LSTM represent an effective approach for speaker- and noise-independent speech enhancement.

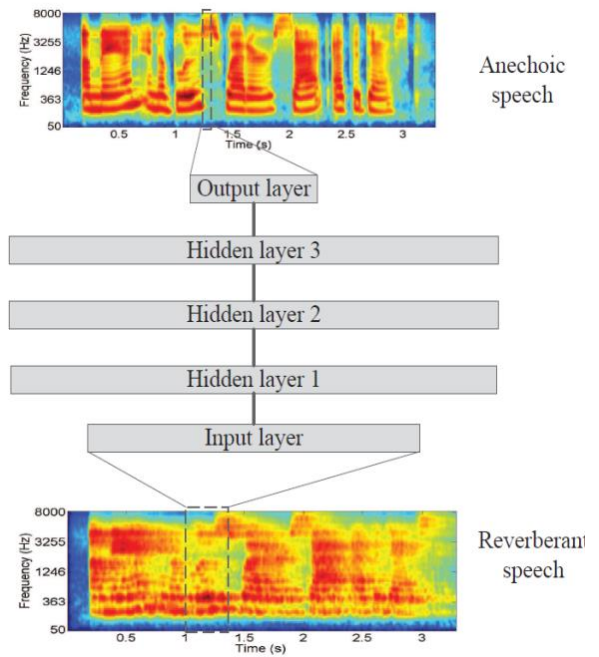

Figure 10. Diagram of a DNN for speech dereverberation based on spectral mapping (from [57]).

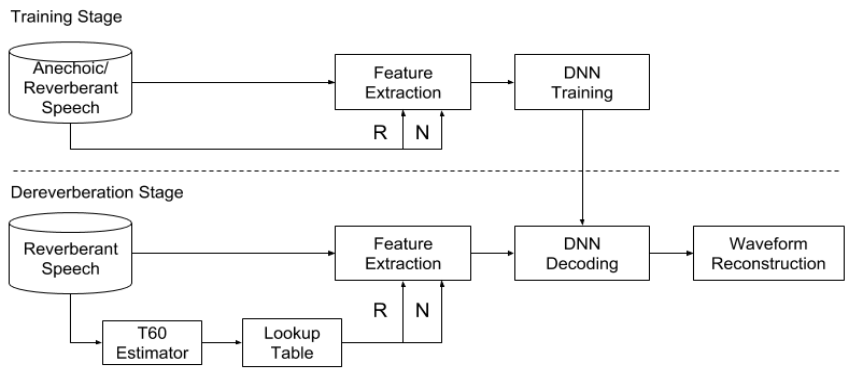

Figure 11. Diagram of a reverberation time aware DNN for speech dereverberation (redrawn from [190]).

## C. Speech Dereverberation and Denoising

In a real environment, speech is usually corrupted by reverberation from surface reflections. Room reverberation corresponds to a convolution of the direct signal and an RIR, and it distorts speech signals along both time and frequency. Reverberation is a well-recognized challenge in speech processing, particularly when it is combined with background noise. As a result, dereverberation has been actively investigated for a long time [5] [191] [131] [61].

Han et al. [57] proposed the first DNN based approach to speech dereverberation. This approach uses spectral mapping on a cochleagram. In other words, a DNN is trained to map from a window of reverberant speech frames to a frame of anechoic speech, as illustrated in Fig. 10. The trained DNN can reconstruct the cochleagram of anechoic speech with surprisingly high quality. In their later work [58], they apply spectral mapping on a spectrogram and extend the approach to perform both dereverberation and denoising.

A more sophisticated system was proposed recently by Wu et al. [190], who observe that dereverberation performance improves when frame length and shift are chosen differently depending on the reverberation time (T60). Based on this observation, their system includes T60 as a control parameter in feature extraction and DNN training. During the dereverberation stage, T60 is estimated and used to choose appropriate frame length and shift for feature extraction. This so-called reverberation-time-aware model is illustrated in Fig. 11. Their comparisons show an improvement in dereverberation performance over the DNN in [58].

To improve the estimation of anechoic speech from reverberant and noisy speech, Xiao et al. [194] proposed a DNN trained to predict static, delta and acceleration features at the same time. The static features are log magnitudes of clean speech, and the delta and acceleration features are derived from the static features. It is argued that DNN that predicts static features well should also predict delta and acceleration features well. The incorporation of dynamic features in the DNN structure helps to improve the estimation of static features for dereverberation.

Zhao et al. [211] observe that spectral mapping is more effective for dereverberation than T-F masking, whereas masking works better than mapping for denoising. Consequently, they construct a two-stage DNN where the first stage performs ratio masking for denoising and the second stage spectral mapping for dereverberation. Furthermore, to alleviate the adverse effects of using the phase of reverberant-noisy speech in resynthesizing the waveform signal of enhanced speech, this study extends the time-domain signal reconstruction technique in [182]. Here the training target is defined in the time-domain, but clean phase is used during training unlike in [182] where noisy phase is used. The two stages are individually trained first, and then jointly trained. The results in [211] show that the two-stage DNN model significantly outperforms the single-stage models for either mapping or masking.

## D. Speaker Separation

The goal of speaker separation is to extract multiple speech signals, one for each speaker, from a mixture containing two or more voices. After deep learning was demonstrated to be capable of speech enhancement, DNN has been successfully applied to speaker separation under a similar framework, which is illustrated in Figure 12 in the case of two-speaker or cochannel separation.

According to our literature search, Huang et al. [81] were the first to introduce DNN for this task. This study addresses two-speaker separation using both a feedforward DNN and an RNN. The authors argue that the summation of the spectra of two estimated sources at frame $t$, $\widehat{S}_1(t)$ and $\widehat{S}_2(t)$, is not guaranteed to equal the spectrum of the mixture. Therefore, a masking layer is added to the network, which produces two final outputs shown in the following equations:

$$\tilde{S}_1(t) = \frac{|\widehat{S}_1(t)|}{|\widehat{S}_1(t)| + |\widehat{S}_2(t)|} \odot Y(t) \qquad (8)$$

$$\tilde{S}_2(t) = \frac{|\widehat{S}_2(t)|}{|\widehat{S}_1(t)| + |\widehat{S}_2(t)|} \odot Y(t) \qquad (9)$$

where $Y(t)$ denotes the mixture spectrum at $t$. This amounts to a signal approximation training target introduced in Section III.I. Both binary and ratio masking are found to be effective. In addition, discriminative training is applied to maximize the difference between one speaker and the estimated version of the other. During training, the following cost is minimized:

$$\frac{1}{2}\sum_t (\|S_1(t) - \tilde{S}_1(t)\|^2 + \|S_2(t) - \tilde{S}_2(t)\|^2 \\ - \gamma \|S_1(t) - \tilde{S}_2(t)\|^2 \\ - \gamma \|S_2(t) - \tilde{S}_1(t)\|^2) \quad (10)$$

where $S_1(t)$ and $S_2(t)$ denote the ground truth spectra for Speaker 1 and Speaker 2, respectively, and $\gamma$ is a tunable parameter. Experimental results have shown that both the masking layer and discriminative training improve speaker separation [82].

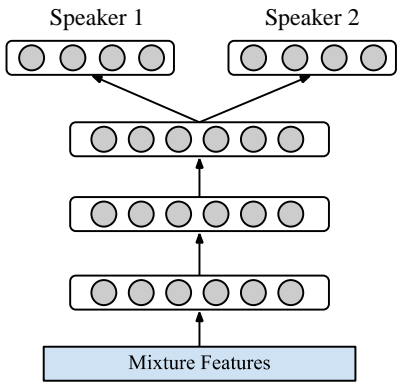

Figure 12. Diagram of DNN based two-speaker separation.

A few months later, Du et al. [38] appeared to have independently proposed a DNN for speaker separation similar to [81]. In this study [38], the DNN is trained to estimate the log power spectrum of the target speaker from that of a cochannel mixture. In a different paper [162], they trained a DNN to map a cochannel signal to the spectrum of the target speaker as well as the spectrum of an interfering speaker, as illustrated in Fig. 12 (see [37] for an extended version). A notable extension compared to [81] is that these papers also address the situation where only the target speaker is the same between training and testing, while interfering speakers are different between training and testing.

In speaker separation, if the underlying speakers are not allowed to change from training to testing, this is the *speaker-dependent* situation. If interfering speakers are allowed to change, but the target speaker is fixed, this is called *target-dependent* speaker separation. In the least constrained case where none of the speakers are required to be the same between training and testing, this is called *speaker-independent*. From this perspective, Huang et al.'s approach is speaker dependent [81] [82] and the studies in [38] [162] deal with both speaker and target dependent separation. Their way of relaxing the constraint on interfering speakers is simply to train with cochannel mixtures of the target speaker and many interferers.

Zhang and Wang proposed a deep ensemble network to address speaker-dependent as well as target-dependent separation [206]. They employ multi-context networks to integrate temporal information at different resolutions. An ensemble is constructed by stacking multiple modules, each performing multi-context masking or mapping. Several training targets were examined in this study. For speaker-dependent separation, signal approximation is shown to be most effective; for target-dependent separation, a combination of ratio masking and signal approximation is most effective. Furthermore, the performance of target-dependent separation is close to that of speaker-dependent separation. Recently, Wang et al. [174] took a step further towards relaxing speaker dependency in talker separation. Their approach clusters each speaker into one of the four clusters (two for male and two for female), and then trains a DNN-based gender mixture detector to determine the clusters of the two underlying speakers in a mixture. Although trained on a subset of speakers in each cluster, their evaluation results show that the speaker separation approach works well for the other untrained speakers in each cluster; in other words, this speaker separation approach exhibits a degree of speaker independency.

Healy et al. [63] have recently used a DNN for speaker-dependent cochannel separation and performed speech intelligibility evaluation of the DNN with both HI and NH listeners. The DNN was trained to estimate the IRM and its complement, corresponding to the target talker and interfering talker. Compared to earlier DNN-based cochannel separation studies, the algorithm in [63] uses a diverse set of features and predicts multiple IRM frames, resulting in better separation. The intelligibility results are shown in Figure 13. For the HI group, intelligibility improvement from DNN-based separation is 42.5, 49.2, and 58.7 percentage points at -3 dB, -6 dB, and -9 dB target-to-interferer ratio (TIR), respectively. For the NH group, there are statistically significant improvements, but to a smaller extent. It is remarkable that the large intelligibility improvements obtained by HI listeners allow them to perform equivalently to NH listeners (without algorithm help) at the common TIRs of -6 and -9 dB.

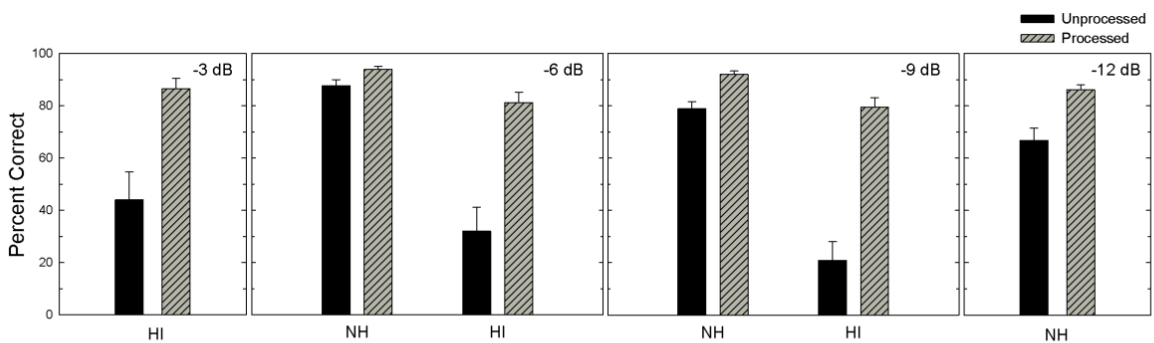

Figure 13. Mean intelligibility scores and standard errors for HI and NH subjects listening to target sentences mixed with interfering sentences and separated target sentences (from [63]). Percent correct results are given at four different target-to-interferer ratios.

Speaker-independent separation can be treated as unsupervised clustering where T-F units are clustered into distinct classes dominated by individual speakers [6] [79]. Clustering is a flexible framework in terms of the number of speakers to separate, but it does not benefit as much from discriminative information fully utilized in supervised training. Hershey et al. were the first to address speaker-independent multi-talker separation in the DNN framework [69]. Their approach, called deep clustering, combines DNN based feature learning and spectral clustering. With a ground truth partition of T-F units, the affinity matrix $A$ can be computed as:

$$A = YY^T \qquad (11)$$

where $Y$ is the indicator matrix built from the IBM. $Y_{i,c}$ is set to 1 if unit $i$ belongs to (or dominated by) speaker $c$, and 0 otherwise. The DNN is trained to embed each T-F unit. The estimated affinity matrix $\widehat{A}$ can be derived from the embeddings. The DNN learns to output similar embeddings for T-F units originating from the same speaker by minimizing the following cost function:

$$C_Y(V) = \left\|\widehat{A} - A\right\|_F^2 = \|VV^T - YY^T\|_F^2 \qquad (12)$$

where $V$ is an embedding matrix for T-F units. Each row of $V$ represents one T-F unit. $\|\cdot\|_F^2$ denotes the squared Frobenius norm. Low rank formulation can be applied to efficiently calculate the cost function and its derivatives. During inference, a mixture is segmented and the embedding matrix $V$ is computed for each segment. Then, the embedding matrices of all segments are concatenated. Finally, the $K$-means algorithm is applied to cluster the T-F units of all the segments into speaker clusters. Segment-level clustering is more accurate than utterance-level clustering, but with clustering results only for individual segments, the problem of sequential organization has to be addressed. Deep clustering is shown to produce high quality speaker separation, significantly better than a CASA method [79] and an NMF method for speaker-independent separation.

A recent extension of deep clustering is the deep attractor network [25], which also learns high-dimensional embeddings for T-F units. Unlike deep clustering, this deep network creates attractor points akin to cluster centers in order to pull T-F units dominated by different speakers to their corresponding attractors. Speaker separation is then performed as mask estimation by comparing embedded points and each attractor. The results in [25] show that the deep attractor network yields better results than deep clustering.

While clustering-based methods naturally lead to speaker-independent models, DNN based masking/mapping methods tie each output of the DNN to a specific speaker, and lead to speaker-dependent models. For example, mapping based methods minimize the following cost function:

$$J = \sum_{k,t} \left\| |\widetilde{S}_k(t)| - |S_k(t)| \right\|^2 \qquad (13)$$

where $|\widetilde{S}_k(t)|$ and $|S_k(t)|$ denote estimated and actual spectral magnitudes for speaker $k$, respectively, and $t$ denotes time frame. To untie DNN outputs from speakers and train a speaker-independent model using a masking or mapping technique, Yu et al. [202] recently proposed permutation-invariant training, which is shown in Fig. 14. For two-speaker separation, a DNN is trained to output two masks, each of which is applied to noisy speech to produce a source estimate. During DNN training, the cost function is dynamically calculated. If we assign each output to a reference speaker $|S_k(t)|$ in the training data, there are two possible assignments, each of which is associated with an MSE. The assignment with the lower MSE is chosen and the DNN is trained to minimize the corresponding MSE. During both training and inference, the DNN takes a segment or multiple frames of features, and estimates two sources for the segment. Since the two outputs of the DNN are not tied to any speaker, the same speaker may switch from one output to another across consecutive segments. Therefore, the estimated segment-level sources need to be sequentially organized unless segments are as long as utterances. Although much simpler, speaker separation results are shown to match those obtained with deep clustering [202] [101].

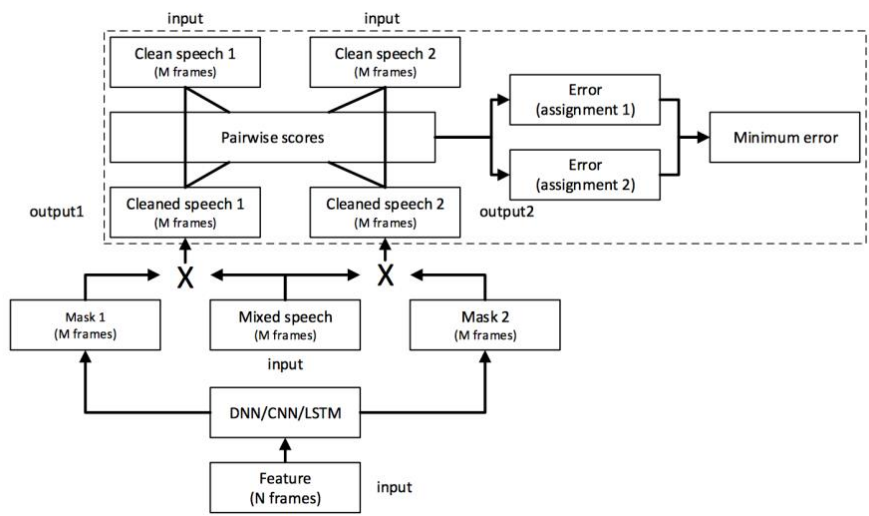

Figure 14. Two-talker separation with permutation-invariant training (from [202]).

It should be noted that, although speaker separation evaluations typically focus on two-speaker mixtures, the separation framework can be generalized to separating more than two talkers. For example, the diagrams in both Figs. 12 and 14 can be straightforwardly extended to handle, say, three-talker mixtures. One can also train target-independent models using multi-speaker mixtures. For speaker-independent separation, deep clustering [69] and permutation-invariant training [101] are both formulated for multi-talker mixtures and evaluated on such data. Scaling deep clustering from mixtures of two speakers to more than two is more straightforward than for scaling permutation-invariant training.

An insight from the body of work overviewed in this speaker separation subsection is that a DNN model trained with many pairs of different speakers is able to separate a pair of speakers never included in training, a case of speaker independent separation, but only at the frame level. For speaker-independent separation, the key issue is how to group well-separated speech signals at individual frames (or segments) across time. This is precisely the issue of sequential organization, which is much investigated in CASA [172]. Permutation-invariant training may be considered as imposing sequential grouping constraints during DNN training. On the other hand, typical CASA methods utilize pitch contours, vocal tract characteristics, rhythm or prosody, and even common spatial direction when multiple sensors are available, which do not usually involve supervised learning. It seems to us that integrating traditional CASA techniques and deep learning is a fertile ground for future research.

## VI. ARRAY SEPARATION ALGORITHMS

An array of microphones provides multiple monaural recordings, which contain information indicative of the spatial origin of a sound source. When sound sources are spatially separated, with sensor array inputs one may localize sound sources and then extract the source from the target location or direction. Traditional approaches to source separation based on spatial information include beamforming, as mentioned in Sect. I, and independent component analysis [8] [85] [3]. Sound localization and location-based grouping are among the classic topics in auditory perception and CASA [12] [15] [172].

### A. Separation Based on Spatial Feature Extraction

The first study in supervised speech segregation was conducted by Roman et al. [141] in the binaural domain. This study performs supervised classification to estimate the IBM based on two binaural features: ITD and ILD, both extracted from individual T-F unit pairs from the left-ear and right-ear cochleagram. Note that, in this case, the IBM is defined on the noisy speech at a single ear (reference channel). Classification is based on maximum *a posteriori* (MAP) probability where the likelihood is given by a density estimation technique. Another classic two-sensor separation technique, DUET (Degenerate Unmixing Estimation Technique), was published by Yilmaz and Rickard [199] at about the same time. DUET is based on unsupervised clustering, and the spatial features used are phase and amplitude differences between the two microphones. The contrast between classification and clustering in these studies is a persistent theme and anticipates similar contrasts in later studies, e.g. binary masking [71] vs. clustering [72] for beamforming (see Sect. VI.B), and deep clustering [69] versus mask estimation [101] for talker-independent speaker separation (see Sect. V.D).

The use of spatial information afforded by an array as features in deep learning is a straightforward extension of the earlier use of DNN in monaural separation; one simply substitutes spatial features for monaural features. Indeed, this way of leveraging spatial information provides a natural framework for integrating monaural and spatial features for source separation, which is a point worth emphasizing as traditional research tends to pursue array separation without considering monaural grouping. It is worth noting that human auditory scene analysis integrates monaural and binaural analysis in a seamless fashion, taking advantage of whatever discriminant information existing in a particular environment [15] [172] [30].

The first study to employ DNN for binaural separation was published by Jiang et al. [90]. In this study, the signals from two ears (or microphones) are passed to two corresponding auditory filterbanks. ITD and ILD features are extracted from T-F unit pairs and sent to a subband DNN for IBM estimation, one DNN for each frequency channel. In addition, a monaural feature (GFCC, see Table 1) is extracted from the left-ear input. A number of conclusions can be drawn from this study. Perhaps most important is the observation that the trained

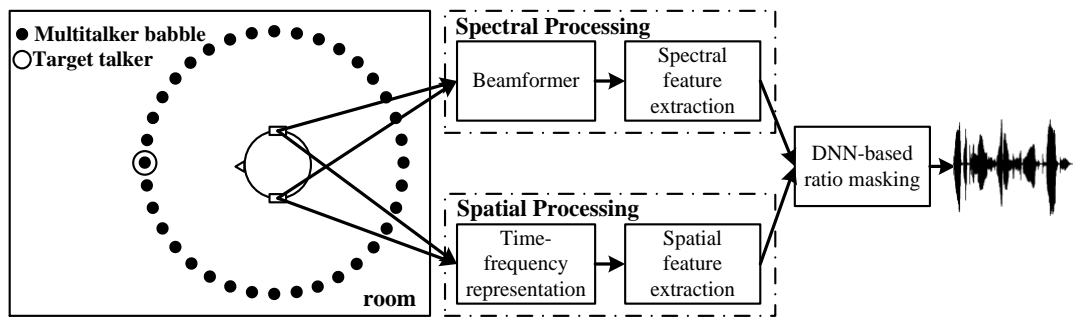

Figure 15. Schematic diagram of a binaural separation algorithm (from [208]).

DNN generalizes well to untrained spatial configurations of sound sources. A spatial configuration refers to a specific placement of sound sources and sensors in an acoustic environment. This is key to the use of supervised learning as there are infinite configurations and a training set cannot enumerate various configurations. DNN based binaural separation is found to generalize well to RIRs and reverberation times. It is also observed that the incorporation of the monaural feature improves separation performance, especially when the target and interfering sources are co-located or close to each other.

Araki et al. [2] subsequently employed a DNN for spectral mapping that includes the spatial features of ILD, interaural phase difference (IPD), and enhanced features with an initial mask derived from location information, in addition to monaural input. Their evaluation with ASR related metrics shows that the best enhancement performance is obtained with a combination of monaural and enhanced features. Fan et al. [43] proposed a spectral mapping approach utilizing both binaural and monaural inputs. For the binaural features, this study uses subband ILDs, which are found to be more effective than fullband ILDs. These features are then concatenated with the left-ear's frame-level log power spectra to form the input to the DNN, which is trained to map to the spectrum of clean speech. A quantitative comparison with [90] shows that their system produces better PESQ scores for separated speech but similar STOI numbers.

A more sophisticated binaural separation algorithm was proposed by Yu et al. [203]. The spatial features used include IPD, ILD, and a so-called mixing vector that is a form of combined STFT values of a unit pair. The DNN used is a DAE, first trained unsupervisedly as autoencoders that are subsequently stacked into a DNN subject to supervised fine-tuning. Extracted spatial features are first mapped to high-level features indicating spatial directions via unsupervised DAE training. For separation, a classifier is trained to map high-level spatial features to a discretized range of source directions. This algorithm operates over subbands, each covering a block of consecutive frequency channels.

Recently, Zhang and Wang [208] developed a DNN for IRM estimation with a more sophisticated set of spatial and spectral features. Their algorithm is illustrated in Fig. 15, where the left-ear and right-ear inputs are fed to two different modules for spectral (monaural) and spatial (binaural) analysis. Instead of monaural analysis on a single ear [90] [43], spectral analysis in [208] is conducted on the output of a fixed beamformer, which itself removes some background inference, by extracting a complementary set of monaural features (see Sect. IV). For spatial analysis, ITD in the form of a cross-correlation function, and ILD are extracted. The spectral and spatial features are concatenated to form the input to a DNN for IRM estimation at the frame level. This algorithm is shown to produce substantially better separation results in reverberant multisource environments than conventional beamformers, including MVDR (Minimum Variance Distortionless Response) and MWF (Multichannel Wiener Filter). An interesting observation from their analysis is that much of the benefit of using a beamformer prior to spectral feature extraction can be obtained simply by concatenating monaural features from the two ears.

Although the above methods are all binaural, involving two sensors, the extension from two sensors to an array with $N$ sensors, with $N > 2$, is usually straightforward. Take the system in Fig. 15, for instance. With $N$ microphones, spectral feature extraction requires no change as traditional beamformers are already formulated for an arbitrary number of microphones. For spatial feature extraction, the feature space needs to be expanded when more than two sensors are available, either by designating one microphone as a reference for deriving a set of "binaural" features or by considering a matrix of all sensor pairs in a correlation or covariance analysis. The output is a T-F mask or spectral envelope corresponding to target speech, which may be viewed as monaural. Since traditional beamforming with an array also produces a "monaural" output, corresponding to the target source, T-F masking based on spatial features may be considered beamforming or, more accurately, nonlinear beamforming [125] as opposed to traditional beamforming that is linear.

### B. Time-frequency Masking for Beamforming

Beamforming, as the name would suggest, tunes in the signals from a zone of arrival angles centered at a given angle, while tuning out the signals outside the zone. To be applicable, a beamformer needs to know the target direction to steer the beamformer. Such a steering vector is typically supplied by estimating the direction-of-arrival (DOA) of the target source, or more broadly sound localization. In reverberant, multi-source environments, localizing the target sound is far from trivial. It is well recognized in CASA that localization and separation are two closely related functions ([172], Chapter 5). For human audition, evidence suggests that sound localization largely depends on source separation [60] [30].

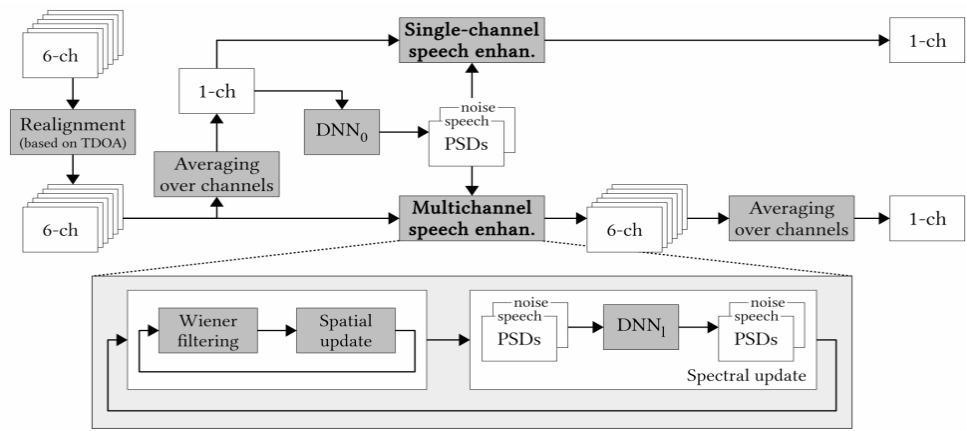

Figure 16. Diagram of a DNN based array source separation method (from [133]).

Fueled by the CHiME-3 challenge for robust ASR, two independent studies made the first use of DNN based monaural speech enhancement in conjunction with conventional beamforming, both published in ICASSP 2016 [71] [72]. The CHiME-3 challenge provides noisy speech data from a single speaker recorded by 6 microphones mounted on a tablet [7]. In these two studies, monaural speech separation provides the basis for computing the steering vector, cleverly bypassing two tasks that would have been required via the DOA estimation: localizing multiple sound sources and selecting the target (speech) source. To explain their idea, let us first describe MVDR as a representative beamformer.

MVDR aims to minimize the noise energy from nontarget directions while imposing linear constraints to maintain the energy from the target direction [45]. The captured signals of an array in the STFT domain can be written as:

$$\mathbf{y}(t,f) = \mathbf{c}(f)s(t,f) + \mathbf{n}(t,f) \quad (14)$$

where $\mathbf{y}(t,f)$ and $\mathbf{n}(t,f)$ denote the STFT spatial vectors of the noisy speech signal and noise at frame $t$ and frequency $f$, respectively, and $s(t,f)$ denotes the STFT of the speech source. The term $\mathbf{c}(f)s(t,f)$ denotes the received speech signal by the array and $\mathbf{c}(f)$ is the steering vector of the array.

At frequency $f$, the MVDR beamformer identifies a weight vector $\mathbf{w}(f)$ that minimizes the average output power of the beamformer while maintaining the energy along the look (target) direction. Omitting $f$ for brevity, this optimization problem can be formulated as

$$\mathbf{w}_{opt} = \underset{\mathbf{w}}{\mathrm{argmin}}\{\mathbf{w}^H \mathbf{\Phi}_n \mathbf{w}\}, \quad \text{subject to } \mathbf{w}^H \mathbf{c} = 1 \quad (15)$$

where $H$ denotes the conjugate transpose and $\mathbf{\Phi}_n$ is the spatial covariance matrix of the noise. Note that the minimization of the output power is equivalent to the minimization of the noise power. The solution to this quadratic optimization problem is:

$$\mathbf{w}_{opt} = \frac{\mathbf{\Phi}_n^{-1}\mathbf{c}}{\mathbf{c}^H \mathbf{\Phi}_n^{-1} \mathbf{c}} \quad (16)$$

The enhanced speech signal is given by

$$\tilde{s}(t) = \mathbf{w}_{opt}^H \mathbf{y}(t) \quad (17)$$

Hence, the accurate estimation of $\mathbf{c}$ and $\mathbf{\Phi}_n$ is key to MVDR beamforming. Furthermore, $\mathbf{c}$ corresponds to the principal component of $\mathbf{\Phi}_x$ (the eigenvector with the largest eigenvalue), the spatial covariance matrix of speech. With speech and noise uncorrelated, we have

$$\mathbf{\Phi}_x = \mathbf{\Phi}_y - \mathbf{\Phi}_n \quad (18)$$

Therefore, a noise estimate is crucial for beamforming performance, just like it is for traditional speech enhancement.

In [71], an RNN with bidirectional LSTM is used for IBM estimation. A common neural network is trained monaurally on the data from each of the sensors. Then the trained network is used to produce a binary mask for each microphone recording, and the multiple masks are combined into one mask with a median operation. The single mask is used to estimate the speech and noise covariance matrix, from which beamformer coefficients are obtained. Their results show that MVDR does not work as well as the GEV (generalized eigenvector) beamformer. In [72], a spatial clustering based approach was proposed to compute a ratio mask. This approach uses a complex-domain GMM (cGMM) to describe the distribution of the T-F units dominated by noise and another cGMM to describe that of the units with both speech and noise. After parameter estimation, the two cGMMs are used for calculating the covariance matrices of noisy speech and noise, which are fed to an MVDR beamformer for speech separation. Both of these algorithms perform very well, and Higuchi et al.'s method was used in the best performing system in the CHiME-3 challenge [200]. A similar approach, i.e. DNN-based IRM estimation combined with a beamformer, is also behind the winning system in the most recent CHiME-4 challenge [36].

A method different from the above two studies was given by Nugraha et al. [133], who perform array source separation using DNN for monaural separation and a complex multivariate Gaussian distribution to model spatial information. The DNN in this study is used to model source spectra, or spectral mapping. The power spectral densities (PSDs) and spatial covariance matrices of speech and noise are estimated and updated iteratively. Figure 16 illustrates the processing pipeline. First, array signals are realigned on the basis of time difference of arrival (TDOA) and averaged to form a monaural signal. A DNN is then used to produce an initial estimate of noise and speech PSDs. During the iterative estimation of PSDs and spatial covariance matrices, DNNs are used to further improve the PSDs estimated by a multichannel

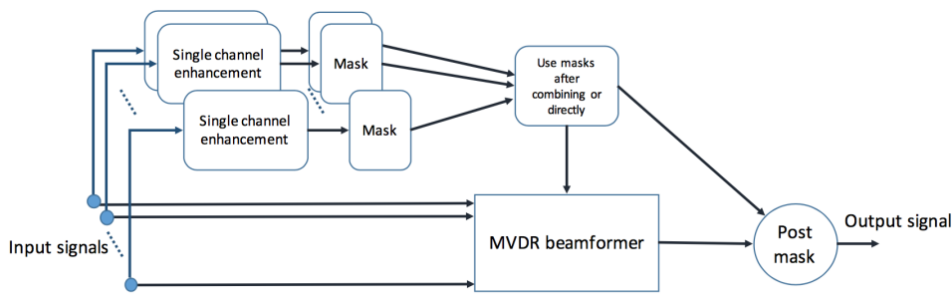

Figure 17. MVDR beamformer with monaural mask estimation (from [42]).

Wiener filter. Finally, the estimated speech signals from multiple microphones are averaged to produce a single speech estimate for ASR evaluation. A number of design choices were examined in this study, and their algorithm yields better separation and ASR results than DNN based monaural separation and an array version of NMF-based separation.

The success of Higuchi et al. [72] and Heymann et al. [71] in the CHiME-3 challenge by using DNN estimated masks for beamforming has motivated many recent studies, exploring different ways of integrating T-F masking and beamforming. Erdogan et al. [42] trained an RNN for monaural speech enhancement, from which a ratio mask is computed in order to provide coefficients for an MVDR beamformer. As illustrated in Fig. 17, a ratio mask is first estimated for each microphone. Then multiple masks from an array are combined into one mask by a maximum operator, which is found to produce better results than using multiple masks without combination. It should be noted that their ASR results on the CHiME-3 data are not compelling. Instead of fixed beamformers like MVDR, beamforming coefficients can be dynamically predicted by a DNN. Li et al. [108] employed a deep network to predict spatial filters from array inputs of noisy speech for adaptive beamforming. Waveform signals are sent to a shared RNN, whose output is sent to two separate RNNs to predict beamforming filters for two microphones.

Zhang et al. [209] trained a DNN for IRM estimation from a complementary set of monaural features, and then combined multiple ratio masks from an array into a single one with a maximum operator. The ratio mask is used for calculating the noise spatial covariance matrix at time $t$ for an MVDR beamformer as follows,

$$\Phi_n(t,f) = \frac{1}{\sum_{l=t-L}^{t+L}(1-RM(l,f))} \times \sum_{l=t-L}^{t+L}(1-RM(l,f))\mathbf{y}(l,f)\mathbf{y}(l,f)^H \quad (19)$$

where $RM(l,f)$ denotes the estimated IRM from the DNN at frame $l$ and frequency $f$. An element of the noise covariance matrix is calculated per frame by integrating a window of neighboring $2L+1$ frames. They find this adaptive way of estimating the noise covariance matrix to perform much better than estimation over the entire utterance or a signal segment. An enhanced speech signal from the beamformer is then fed to the DNN to refine the IRM estimate, and mask estimation and beamforming iterate several times to produce the final output. Their 5.05 WER (word error rate) on the CHiME-3 real evaluation data represents a 13.34% relative improvement over the previous best [200]. Independently, Xiao et al. [193] also proposed to iterate ratio masking and beamforming. They use an RNN for estimating a speech mask and a noise mask. Mask refinement is based on an ASR loss, in order to directly benefit ASR performance. They showed that this approach leads to a considerable WER reduction over the use of a conventional MVDR, although recognition accuracy is not as high as in [200].

Other related studies include Pfeifenberger et al. [139], who use the cosine distance between the principal components of consecutive frames of noisy speech as the feature for DNN mask estimation. Meng et al. [121] use RNNs for adaptive estimation of beamformer coefficients. Their ASR results on the CHiME-3 data are better than the baseline scores, but are far from the best scores. Nakatani et al. [129] integrate DNN mask estimation and cGMM clustering based estimation to further improve the quality of mask estimates. Their results on the CHiME-3 data improve over those obtained from RNN or cGMM generated masks.

## VII. DISCUSSION AND CONCLUSION

This paper has provided a comprehensive overview of DNN based supervised speech separation. We have summarized key components of supervised separation, i.e. learning machines, training targets, and acoustic features, explained representative algorithms, and reviewed a large number of related studies. With the formulation of the separation problem as supervised learning, DNN based separation over a short few years has greatly elevated the state-of-the-art for a wide range of speech separation tasks, including monaural speech enhancement, speech dereverberation, and speaker separation, as well as array speech separation. This rapid advance will likely continue with a tighter integration of domain knowledge and the data-driven framework and the progress in deep learning itself.

Below we discuss several conceptual issues pertinent to this overview.

### A. Features vs. Learning Machines

As discussed in Sect. IV, features are important for speech separation. However, a main appeal of deep learning is to *learn* appropriate features for a task, rather than to *design* such features. So is there a role for feature extraction in the era of deep learning? We believe the answer is yes. The so-called no-free-lunch theorem [189] dictates that no learning algorithm, DNN included, achieves superior performance in all tasks. Aside from theoretical arguments, feature extraction is a way of imparting knowledge from a problem domain and it stands to reason that it is useful to incorporate domain knowledge this way (see [176] for a recent example). For instance, the success of CNN in visual pattern recognition is partly due to the use of shared weights and pooling (sampling) layers in its architecture that helps to build a representation invariant to small variations of feature positions [10].

It is possible to learn useful features for a problem domain, but doing so may not be computationally efficient, particularly when certain features are known to be discriminative through domain research. Take pitch, for example. Much research in auditory scene analysis shows that pitch is a primary cue for auditory organization [15] [30], and research in CASA demonstrates that pitch alone can go a long way in separating voiced speech [78]. Perhaps a DNN can be trained to "discover" harmonicity as a prominent feature, and there is some hint at this from a recent study [24], but extracting pitch as input features seems like the most straightforward way of incorporating pitch in speech separation.

The above discussion is not meant to discount the importance of learning machines, as this overview has made it abundantly clear, but to argue for the relevance of feature extraction despite the power of deep learning. As mentioned in Sect. V.A, convolutional layers in a CNN amount to feature extraction. Although CNN weights are trained, the use of a particular CNN architecture reflects design choices of its user.

*B.     Time-frequency Domain vs. Time Domain*

The vast majority of supervised speech separation studies are conducted in the T-F domain as reflected in the various training targets reviewed in Sect. III. Alternatively, speech separation can be conducted in the time domain without recourse to a frequency representation. As pointed out in Sect. V.A, through temporal mapping both magnitude and phase can potentially be cleaned at once. End-to-end separation represents an emergent trend along with the use of CNNs and GANs.

A few comments are in order. First, temporal mapping is a welcome addition to the list of supervised separation approaches and provides a unique perspective to phase enhancement [50] [103]. Second, the same signal can be transformed back and forth between its time domain representation and its T-F domain representation. Third, the human auditory system has a frequency dimension at the beginning of the auditory pathway, i.e. at the cochlea. It is interesting to note Licklider's classic duplex theory of pitch perception, postulating two processes of pitch analysis: a spatial process corresponding to the frequency dimension in the cochlea and a temporal process corresponding to the temporal response of each frequency channel [111]. Computational models for pitch estimation fall into three categories: spectral, temporal, and spectrotemporal approaches [33]. In this sense, a cochleagram, with the individual responses of a cochlear filterbank [118] [172], is a duplex representation.

*C.     What's the Target?*

When multiple sounds are present in the acoustic environment, which should be treated as the target sound at a particular time? The definition of ideal masks presumes that the target source is known, which is often the case in speech separation applications. For speech enhancement, the speech signal is considered the target while nonspeech signals are considered the interference. The situation becomes tricky for multi-speaker separation. In general, this is the issue of auditory attention and intention. It is a complicated issue as what is attended to shifts from one moment to the next even with the same input scene, and does not have to be a speech signal. There are, however, practical solutions. For example, directional hearing aids get around this issue by assuming that the target lies in the look direction, i.e. benefiting from vision [170] [35]. With sources separated, there are other reasonable alternatives for target definition, e.g. the loudest source, the previously attended one (i.e. tracking), or the most familiar (as in the multi-speaker case). A full account, however, would require a sophistical model of auditory attention (see [172] [118]).

*D.     What Does a Solution to the Cocktail Party Problem Look Like?*

CASA defines the solution to the cocktail party problem as a system that achieves human separation performance in *all listening conditions* ([172], p.28). But how to actually compare the separation performance by a machine and that by a human listener? Perhaps a straightforward way would be compare ASR scores and human speech intelligibility scores in various listening conditions. This is a tall order as ASR performance still falls short in realistic conditions despite tremendous recent advances thanks to deep learning. A drawback with ASR evaluation is the dependency on ASR with all its peculiarities.

Here we suggest a different, concrete measure: a solution to the cocktail party is *a separation system that elevates speech intelligibility of hearing-impaired listeners to the level of normal-hearing listeners in all listening situations*. Not as broad as defined in CASA, but this definition has the benefit that it is tightly linked to a primary driver for speech separation research, namely, to eliminate the speech understanding handicap of millions of listeners with impaired hearing [171]. By this definition, the DNN based speech enhancement described above has met the criterion in limited conditions (see Fig. 13 for one example), but clearly not in all conditions. Versatility is the hallmark of human intelligence, and the primary challenge facing supervised speech separation research today.

Before closing, we point out that the use of supervised learning and DNN in signal processing goes beyond speech separation, and automatic speech and speaker recognition. The related topics include multipitch tracking [80] [56], voice activity detection [207], and even a task as basic in signal processing as SNR estimation [134]. No matter the task, once it is formulated as a data-driven problem, advances will likely ensue with the use of various deep learning models and suitably constructed training sets; it should also be mentioned that these advances come at the expense of high computational complexity involved in the training process and often in operating a trained DNN model. A considerable benefit of treating signal processing as learning is that signal processing can ride on the progress of machine learning, a rapidly advancing field.

Finally, we remark that human ability to solve the cocktail party problem appears to have much to do with our extensive exposure to various noisy environments (see also [24]). Research indicates that children have poorer ability to recognize speech in noise than adults [54] [92], and musicians are better at perceiving noisy speech than non-musicians [135] presumably due to musicians' long exposure to polyphonic signals. Relative to monolingual speakers, bilinguals have a deficit when it comes to speech perception in noise, although the two groups are similarly proficient in quiet [159]. All these effects support the notion that extensive training (experience) is part of the reason for the remarkable robustness of the normal auditory system to acoustic interference.

**Acknowledgments**: The preparation of this overview was supported in part by an AFOSR grant (FA9550-12-1-0130), an NIDCD grant (R01 DC012048), and an NSF grant (IIS-1409431). We thank Masood Delfarah for help in manuscript preparation and Jun Du, Yu Tsao, Yuxuan Wang, Yong Xu, and Xueliang Zhang for helpful comments on an earlier version.

# REFERENCES

[1] M.C. Anzalone, L. Calandruccio, K.A. Doherty, and L.H. Carney, "Determination of the potential benefit of time-frequency gain manipulation," *Ear Hear.*, vol. 27, pp. 480-492, 2006.

[2] S. Araki, *et al.*, "Exploring multi-channel features for denoising-autoencoder-based speech enhancement," in *Proceedings of ICASSP*, pp. 116-120, 2015.

[3] S. Araki, H. Sawada, R. Mukai, and S. Makino, "Underdetermined blind sparse source separation for arbitrarily arranged multiple sensors," *Sig. Proc.*, vol. 87, pp. 1833-1847, 2007.

[4] P. Assmann and A.Q. Summerfield, "The perception of speech under adverse conditions," in *Speech processing in the auditory system,* S. Greenberg, W.A. Ainsworth, A.N. Popper, and R.R. Fay, Ed., New York: Springer, pp. 231-308, 2004.

[5] C. Avendano and H. Hermansky, "Study on the dereverberation of speech based on temporal envelope filtering," in *Proceedings of ICSLP*, pp. 889-892, 1996.

[6] F.R. Bach and M.I. Jordan, "Learning spectral clustering, with application to speech separation," *J. Mach. Learn. Res.*, vol. 7, pp. 1963-2001, 2006.

[7] J. Barker, R. Marxer, E. Vincent, and A. Watanabe, "The third CHiME speech separation and recognition challenge: dataset, task and baselines," in *Proceedings of IEEE ASRU*, pp. 5210-5214, 2015.

[8] A.J. Bell and T.J. Sejnowski, "An information-maximization approach to blind separation and blind deconvolution," *Neural Comp.*, vol. 7, pp. 1129-1159, 1995.

[9] J. Benesty, J. Chen, and Y. Huang, *Microphone array signal processing.* Berlin: Springer, 2008.

[10] Y. Bengio and Y. LeCun, "Scaling learning algorithms towards AI," in *Large-scale kernel machines,* L. Bottou, O. Chapelle, D. DeCoste, and J. Weston, Ed., Cambridge M.A.: MIT Press, pp. 321-359, 2007.

[11] C. Bey and S. McAdams, "Schema-based processing in auditory scene analysis," *Percept. Psychophys.*, vol. 64, pp. 844-854, 2002.

[12] J. Blauert, *Spatial Hearing: The psychophysics of human sound localization.* Cambridge, MA: MIT Press, 1983.

[13] S.F. Boll, "Suppression of acoustic noise in speech using spectral subtraction," *IEEE Trans. Acoust. Speech Sig. Process.*, vol. 27, pp. 113-120, 1979.

[14] M.S. Brandstein and D.B. Ward, Ed., *Microphone arrays: Signal processing techniques and applications*. New York: Springer, 2001.

[15] A.S. Bregman, *Auditory scene analysis.* Cambridge MA: MIT Press, 1990.

[16] D.S. Brungart, P.S. Chang, B.D. Simpson, and D.L. Wang, "Isolating the energetic component of speech-on-speech masking with ideal time-frequency segregation," *J. Acoust. Soc. Am.*, vol. 120, pp. 4007-4018, 2006.

[17] R.P. Carlyon, R. Cusack, J.M. Foxton, and I.H. Robertson, "Effects of attention and unilateral neglect on auditory stream segregation," *Journal of Experimental Psychology: Human Perception and Performance*, vol. 27, pp. 115-127, 2001.

[18] S.E. Chazan, S. Gannot, and J. Goldberger, "A phoneme-based pre-training approach for deep neural network with application to speech enhancement," in *Proceedings of IWAENC*, 2016.

[19] C. Chen and J.A. Bilmes, "MVA processing of speech features," *IEEE Trans. Audio Speech Lang. Proc.*, vol. 15, pp. 257-270, 2007.

[20] J. Chen and D.L. Wang, "Long short-term memory for speaker generalization in supervised speech separation," in *Proceedings of Interspeech*, pp. 3314-3318, 2016.

[21] J. Chen and D.L. Wang, "DNN-based mask estimation for supervised speech separation," in *Audio source separation,* S. Makino, Ed., Berlin: Springer, pp. 207-235, 2018.

[22] J. Chen, Y. Wang, and D.L. Wang, "A feature study for classification-based speech separation at low signal-to-noise ratios," *IEEE/ACM Trans. Audio Speech Lang. Proc.*, vol. 22, pp. 1993-2002, 2014.

[23] J. Chen, Y. Wang, and D.L. Wang, "Noise perturbation for supervised speech separation," *Speech Comm.*, vol. 78, pp. 1-10, 2016.

[24] J. Chen, Y. Wang, S.E. Yoho, D.L. Wang, and E.W. Healy, "Large-scale training to increase speech intelligibility for hearing-impaired listeners in novel noises," *J. Acoust. Soc. Am.*, vol. 139, pp. 2604-2612, 2016.

[25] Z. Chen, Y. Luo, and N. Mesgarani, "Deep attractor network for single-microphone speaker separation," in *Proceedings of ICASSP.* pp. 246-250, 2017.

[26] E.C. Cherry, "Some experiments on the recognition of speech, with one and with two ears," *J. Acoust. Soc. Am.*, vol. 25, pp. 975-979, 1953.

[27] E.C. Cherry, *On human communication.* Cambridge MA: MIT Press, 1957.

[28] D.C. Ciresan, U. Meier, and J. Schmidhuber, "Multi-column deep neural networks for image classification," in *Proceedings of CVPR*, pp. 3642-3649, 2012.

[29] C.J. Darwin, "Auditory grouping," *Trends in Cognitive Sciences*, vol. 1, pp. 327-333, 1997.

[30] C.J. Darwin, "Listening to speech in the presence of other sounds," *Phil. Trans. Roy. Soc. B*, vol. 363, pp. 1011-1021, 2008.

[31] C.J. Darwin and R.W. Hukin, "Effectiveness of spatial cues, prosody, and talker characteristics in selective attention," *J. Acoust. Soc. Am.*, vol. 107, pp. 970-977, 2000.

[32] M. David, M. Lavandier, N. Grimault, and A. Oxenham, "Sequential stream segregation of voiced and unvoiced speech sounds based on fundamental frequency," *Hearing Research*, vol. 344, pp. 235-243, 2017.

[33] A. de Cheveigne, "Multiple F0 estimation," in *Computational auditory scene analysis: Principles, algorithms, and Applications,* D.L. Wang and G.J. Brown, Ed., Hoboken NJ: Wiley & IEEE Press, pp. 45-79, 2006.

[34] M. Delfarah and D.L. Wang, "Features for masking-based monaural speech separation in reverberant conditions," *IEEE/ACM Trans. Audio Speech Lang. Proc.*, vol. 25, pp. 1085-1094, 2017.

[35] H. Dillon, *Hearing aids.* 2nd ed., Turramurra, Australia: Boomerang, 2012.

[36] J. Du, *et al.*, "The USTC-iFlyteck system for the CHiME4 challenge," in *Proceedings of the CHiME-4 Workshop*, 2016.

[37] J. Du, Y. Tu, L.-R. Dai, and C.-H. Lee, "A regression approach to single-channel speech separation via high-resolution deep neural networks," *IEEE/ACM Trans. Audio Speech Lang. Proc.*, vol. 24, pp. 1424-1437, 2016.

[38] J. Du, Y. Tu, Y. Xu, L.-R. Dai, and C.-H. Lee, "Speech separation of a target speaker based on deep neural networks," in *Proceedings of ICSP*, pp. 65-68, 2014.

[39] J. Du and Y. Xu, "Hierarchical deep neural network for multivariate regresss," *Pattern Recognition*, vol. 63, pp. 149-157, 2017.

[40] Y. Ephraim and D. Malah, "Speech enhancement using a minimum mean-square error short-time spectral amplitude estimator," *IEEE Trans. Acoust. Speech Sig. Process.*, vol. 32, pp. 1109-1121, 1984.

[41] H. Erdogan, J. Hershey, S. Watanabe, and J. Le Roux, "Phase-sensitive and recognition-boosted speech separation

using deep recurrent neural networks," in *Proceedings of ICASSP*, pp. 708-712, 2015.

[42] H. Erdogan, J.R. Hershey, S. Watanabe, M. Mandel, and J.L. Roux, "Improved MVDR beamforming using single-channel mask prediction networks," in *Proceedings of Interspeech*, pp. 1981-1985, 2016.

[43] N. Fan, J. Du, and L.-R. Dai, "A regression approach to binaural speech segregation via deep neural network," in *Proceedings of ISCSLP*, pp. 116-120, 2016.

[44] J.M. Festen and R. Plomp, "Effects of fluctuating noise and interfering speech on the speech-reception threshold for impaired and normal hearing," *J. Acoust. Soc. Am.*, vol. 88, pp. 1725-1736, 1990.

[45] O.L. Frost, "An algorithm for linearly constrained adaptive array processing," *Proc. IEEE*, vol. 60, pp. 926-935, 1972.

[46] S.-W. Fu, Y. Tsao, and X. Lu, "SNR-aware convolutional neural network modeling for speech enhancement," in *Proceedings of Interspeech*, pp. 3678-3772, 2016.

[47] S.-W. Fu, Y. Tsao, X. Lu, and H. Kawai, "Raw waveform-based speech enhancement by fully convolutional networks," *arXiv:1703.02205v3*, 2017.

[48] T. Gao, J. Du, L.-R. Dai, and C.-H. Lee, "SNR-based progressive learning of deep neural network for speech enhancement," in *Proceedings of Interspeech*, pp. 3713-3717, 2016.

[49] E. Gaudrian, N. Grimault, E.W. Healy, and J.-C. Béra, "The relationship between concurrent stream segregation, pitch-based streaming of vowel sequences, and frequency selectivity," *Acta Acoustica United with Acustica*, vol. 98, pp. 317-327, 2012.

[50] T. Gerkmann, M. Krawczyk-Becker, and J. Le Roux, "Phase processing for single-channel speech enhancement: History and recent advances," *IEEE Sig. Proc. Mag.*, vol. 32, pp. 55-66, 2015.

[51] S. Gonzalez and M. Brookes, "Mask-based enhancement for very low quality speech," in *Proceedings of ICASSP*, pp. 7029–7033, 2014.

[52] I.J. Goodfellow, *et al.*, "Generative adversarial nets," in *Proceedings of NIPS*, pp. 2672-2680, 2014.

[53] A. Graves, *et al.*, "A novel connectionist system for unconstrained handwriting recognition," *IEEE Trans. Pattern Anal. Machine Intell.*, vol. 31, pp. 855–868, 2009.

[54] J.W. Hall, J.H. Grose, E. Buss, and M.B. Dev, "Spondee recognition in a two-talker and a speech-shaped noise masker in adults and children," *Ear Hear.*, vol. 23, pp. 159-165, 2002.

[55] K. Han and D.L. Wang, "A classification based approach to speech separation," *J. Acoust. Soc. Am.*, vol. 132, pp. 3475-3483, 2012.

[56] K. Han and D.L. Wang, "Neural network based pitch tracking in very noisy speech," *IEEE/ACM Trans. Audio Speech Lang. Proc.*, vol. 22, pp. 2158-2168, 2014.

[57] K. Han, Y. Wang, and D.L. Wang, "Learning spectral mapping for speech dereverebation," in *Proceedings of ICASSP*, pp. 4661-4665, 2014.

[58] K. Han, *et al.*, "Learning spectral mapping for speech dereverberation and denoising," *IEEE/ACM Trans. Audio Speech Lang. Proc.*, vol. 23, pp. 982-992, 2015.

[59] W. Han, *et al.*, "Perceptual improvement of deep neural networks for monaural speech enhancement," in *Proceedings of IWAENC*, 2016.

[60] W.M. Hartmann, "How we localize sounds," *Physics Today*, 24-29, November 1999.

[61] O. Hazrati, J. Lee, and P.C. Loizou, "Blind binary masking for reverberation suppression in cochlear implants," *J. Acoust. Soc. Am.*, vol. 133, pp. 1607-1614, 2013.

[62] K. He, X. Zhang, S. Ren, and J. Sun, "Deep residual learning for image recognition," in *Proceedings of CVPR*, pp. 770–778, 2016.

[63] E.W. Healy, M. Delfarah, J.L. Vasko, B.L. Carter, and D.L. Wang, "An algorithm to increase intelligibility for hearing-impaired listeners in the presence of a competing talker," *J. Acoust. Soc. Am.*, vol. 141, pp. 4230-4239, 2017.

[64] E.W. Healy, S.E. Yoho, J. Chen, Y. Wang, and D.L. Wang, "An algorithm to increase speech intelligibility for hearing-impaired listeners in novel segments of the same noise type," *J. Acoust. Soc. Am.*, vol. 138, pp. 1660-1669, 2015.

[65] E.W. Healy, S.E. Yoho, Y. Wang, and D.L. Wang, "An algorithm to improve speech recognition in noise for hearing-impaired listeners," *J. Acoust. Soc. Am.*, vol. 134, pp. 3029-3038, 2013.

[66] R.C. Hendriks, R. Heusdens, and J. Jensen, "MMSE based noise PSD tracking with low complexity," in *Proceedings of ICASSP*, pp. 4266-4269, 2010.

[67] H. Hermansky, "Perceptual linear predictive (PLP) analysis of speech," *J. Acoust. Soc. Am.*, vol. 87, pp. 1738-1752, 1990.

[68] H. Hermansky and N. Morgan, "RASTA processing of speech," *IEEE Trans. Speech Audio Proc.*, vol. 2, pp. 578-589, 1994.

[69] J. Hershey, Z. Chen, J. Le Roux, and S. Watanabe, "Deep clustering: Discriminative embeddings for segmentation and separation," in *Proceedings of ICASSP*, pp. 31-35, 2016.

[70] H. Hertz, A. Krogh, and R.G. Palmer, *Introduction to the theory of neural computation.* Redwood City, CA: Addison-Wesley, 1991.

[71] J. Heymann, L. Drude, and R. Haeb-Umbach, "Neural network based spectral mask estimation for acoustic beamforming," in *Proceedings of ICASSP*, pp. 196-200, 2016.

[72] T. Higuchi, N. Ito, T. Yoshioka, and e. al., "Robust MVDR beamforming using time-frequency masks for online/offline ASR in noise," in *Proceedings of ICASSP*, pp. 5210-5214, 2016.

[73] G. Hinton, *et al.*, "Deep neural networks for acoustic modeling in speech recognition," *IEEE Sig. Proc. Mag.*, 82-97, November 2012.

[74] G.E. Hinton, S. Osindero, and Y.-W. Teh, "A fast learning algorithm for deep belief nets," *Neural Comp.*, vol. 18, pp. 1527-1554, 2006.

[75] S. Hochreiter and J. Schmidhuber, "Long short-term memory," *Neural Comp.*, vol. 9, pp. 1735-1780, 1997.

[76] G. Hu and D.L. Wang, "Speech segregation based on pitch tracking and amplitude modulation," in *Proceedings of IEEE WASPAA*, pp. 79-82, 2001.

[77] G. Hu and D.L. Wang, "Monaural speech segregation based on pitch tracking and amplitude modulation," *IEEE Trans. Neural Net.*, vol. 15, pp. 1135-1150, 2004.

[78] G. Hu and D.L. Wang, "A tandem algorithm for pitch estimation and voiced speech segregation," *IEEE Trans. Audio Speech Lang. Proc.*, vol. 18, pp. 2067-2079, 2010.

[79] K. Hu and D.L. Wang, "An unsupervised approach to cochannel speech separation," *IEEE Trans. Audio Speech Lang. Proc.*, vol. 21, pp. 120-129, 2013.

[80] F. Huang and T. Lee, "Pitch estimation in noisy speech using accumulated peak spectrum and sparse estimation technique," *IEEE Trans. Speech Audio Proc.*, vol. 21, pp. 99–109, 2013.

[81] P.-S. Huang, M. Kim, M. Hasegawa-Johnson, and P. Smaragdis, "Deep learning for monaural speech separation," in *Proceedings of ICASSP*, pp. 1581-1585, 2014.

[82] P.-S. Huang, M. Kim, M. Hasegawa-Johnson, and P. Smaragdis, "Joint optimization of masks and deep recurrent neural networks for monaural source separation," *IEEE/ACM*

[83] L. Hui, *et al.*, "Convolutional maxout neural networks for speech separation," in *Proceedings of ISSPIT*, pp. 24-27, 2015.

[84] C. Hummersone, T. Stokes, and T. Brooks, "On the ideal ratio mask as the goal of computational auditory scene analysis," in *Blind Source Separation,* G.R. Naik and W. Wang, Ed., Berlin: Springer, pp. 349-368, 2014.

[85] A. Hyvärinen and E. Oja, "Independent component analysis: Algorithms and applications," *Neural Netw.*, vol. 13, pp. 411-430, 2000.

[86] S. Ikbal, H. Misra, and H.A. Bourlard, "Phase autocorrelation (PAC) derived robust speech features," in *Proceedings of ICASSP*, pp. II.133-136, 2003.

[87] ITU, "Perceptual evaluation of speech quality (PESQ) , and objective method for end-to-end speech quality assessment of narrowband telephone networks and speech codecs," *ITU-T Recommendation P. 862*, 2000.

[88] D.P. Jarrett, E. Habets, and P.A. Naylor, *Theory and applications of spherical microphone array processing.* Switzerland: Springer, 2016.

[89] J. Jensen and C.H. Taal, "An algorithm for predicting the intelligibility of speech masked by modulated noise maskers," *IEEE/ACM Trans. Audio Speech Lang. Proc.*, vol. 24, pp. 2009-2022, 2016.

[90] Y. Jiang, D.L. Wang, R.S. Liu, and Z.M. Feng, "Binaural classification for reverberant speech segregation using deep neural networks," *IEEE/ACM Trans. Audio Speech Lang. Proc.*, vol. 22, pp. 2112-2121, 2014.

[91] Z. Jin and D.L. Wang, "A supervised learning approach to monaural segregation of reverberant speech," *IEEE Trans. Audio Speech Lang. Proc.*, vol. 17, pp. 625-638, 2009.

[92] P.M. Johnstone and R.Y. Litovsky, "Effect of masker type and age on speech intelligibility and spatial release from masking in children and adults," *J. Acoust. Soc. Am.*, vol. 120, pp. 2177–2189, 2006.

[93] G. Kidd, *et al.*, "Determining the energetic and informational components of speech-on-speech masking," *J. Acoust. Soc. Am.*, vol. 140, pp. 132-144, 2016.

[94] C. Kim and R.M. Stern, "Nonlinear enhancement of onset for robust speech recognition," in *Proceedings of Interspeech*. pp. 2058-2061, 2010.

[95] C. Kim and R.M. Stern, "Power-normalized cepstral coefficients (PNCC) for robust speech recognition," in *Proceedings of ICASSP*, pp. 4101-4104, 2012.

[96] D. Kim, S.H. Lee, and R.M. Kil, "Auditory processing of speech signals for robust speech recognition in real-world noisy environments," *IEEE Trans. Speech Audio Proc.*, vol. 7, pp. 55-69, 1999.

[97] G. Kim, Y. Lu, Y. Hu, and P.C. Loizou, "An algorithm that improves speech intelligibility in noise for normal-hearing listeners," *J. Acoust. Soc. Am.*, vol. 126, pp. 1486-1494, 2009.

[98] M. Kim and P. Smaragdis, "Adaptive denoising autoencoders: a fine-tuning scheme to learn from test mixtures," in *Proceedings of LVA/ICA*, pp. 100-107, 2015.

[99] U. Kjems, J.B. Boldt, M.S. Pedersen, T. Lunner, and D.L. Wang, "Role of mask pattern in intelligibility of ideal binary-masked noisy speech," *J. Acoust. Soc. Am.*, vol. 126, pp. 1415-1426, 2009.

[100] M. Kolbak, Z.H. Tan, and J. Jensen, "Speech intelligibility potential of general and specialized deep neural network based speech enhancement systems," *IEEE/ACM Trans. Audio Speech Lang. Proc.*, vol. 25, pp. 153-167, 2017.

[101] M. Kolbak, D. Yu, Z.-H. Tan, and J. Jensen, "Multi-talker speech separation with utterance-level permutation invariant training of deep recurrent neural networks " *IEEE/ACM Trans. Audio Speech Lang. Proc.*, vol. 25, pp. 1901-1913, 2017.

[102] A.A. Kressner, T. May, and C.J. Rozell, "Outcome measures based on classification performance fail to predict the intelligibility of binary-masked speech," *J. Acoust. Soc. Am.*, vol. 139, pp. 3033-3036, 2016.

[103] J. Kulmer and P. Mowlaee, "Phase estimation in single channel speech enhancement using phase decomposition," *IEEE Sig. Proc. Lett.*, vol. 22, pp. 598-602, 2014.

[104] K. Kumar, C. Kim, and R.M. Stern, "Delta-spectral cepstral coefficients for robust speech recognition," in *Proceedings of ICASSP*. pp. 4784-4787, 2011.

[105] J. Le Roux, J. Hershey, and F. Weninger, "Deep NMF for speech separation," in *Proceedings of ICASSP*, pp. 66-70, 2015.

[106] Y. LeCun, *et al.*, "Backpropagation applied to handwritten zip code recognition," *Neural Comp.*, vol. 1, pp. 541-551, 1989.

[107] Y.-S. Lee, C.-Y. Yang, S.-F. Wang, J.-C. Wang, and C.-H. Wu, "Fully complex deep neural network for phase-incorporating monaural source separation," in *Proceedings of ICASSP*, pp. 281-285, 2017.

[108] B. Li, T.N. Sainath, R.J. Weiss, K.W. Wilson, and M. Bacchiani, "Neural network adaptive beamforming for robust multichannel speech recognition " in *Proceedings of Interspeech*, pp. 1976-1980, 2016.

[109] N. Li and P.C. Loizou, "Factors influencing intelligibility of ideal binary-masked speech: Implications for noise reduction," *J. Acoust. Soc. Am.*, vol. 123, pp. 1673-1682, 2008.

[110] S. Liang, W. Liu, W. Jiang, and W. Xue, "The optimal ratio time-frequency mask for speech separation in terms of the signal-to-noise ratio," *J. Acoust. Soc. Am.*, vol. 134, pp. 452-458, 2013.

[111] J.C.R. Licklider, "A duplex theory of pitch perception," *Experientia*, vol. 7, pp. 128-134, 1951.

[112] L. Lightburn and M. Brookes, "SOBM - a binary mask for noisy speech that optimises an objective intelligibility metric," in *Proceedings of ICASSP*, pp. 661-665, 2015.

[113] P.C. Loizou, *Speech enhancement: Theory and practice.* 2nd ed., Boca Raton FL: CRC Press, 2013.

[114] P.C. Loizou and G. Kim, "Reasons why current speech-enhancement algorithms do not improve speech intelligibility and suggested solutions," *IEEE Trans. Audio Speech Lang. Proc.*, vol. 19, pp. 47-56, 2011.

[115] X. Lu, S. Matsuda, C. Hori, and H. Kashioka, "Speech restoration based on deep learning autoencoder with layer-wised pretraining," in *Proceedings of Interspeech*, pp. 1504-1507, 2012.

[116] X. Lu, Y. Tsao, S. Matsuda, and C. Hori, "Speech enhancement based on deep denoising autoencoder," in *Proceedings of Interspeech*, pp. 555-559, 2013.

[117] R.F. Lyon, "A computational model of binaural localization and separation," in *Proceedings of ICASSP*, pp. 1148-1151, 1983.

[118] R.F. Lyon, *Human and machine hearing.* New York: Cambridge University Press, 2017.

[119] H.K. Maganti and M. Matassoni, "An auditory based modulation spectral feature for reverberant speech recognition," in *Proceedings of Interspeech*. pp. 570-573, 2010.

[120] K. Masutomi, N. Barascud, M. Kashino, J.H. McDermott, and M. Chait, "Sound segregation via embedded repetition is robust to inattention," *Journal of Experimenal Psychology: Human Perception and Performance*, vol. 42, pp. 386-400, 2016.

[121] Z. Meng, S. Watanabe, J. Hershey, and H. Erdogan, "Deep long short-term memory adaptive beamforming networks for multichannel robust speech recognition," in *Proceedings of ICASSP*, pp. 271-275, 2017.

[122] D. Michelsanti and Z.-H. Tan, "Conditional generative adversarial networks for speech enhancement and noise-robust speaker verification," in *Proceedings of Interspeech*. pp. 2008-2012, 2017.

[123] G.A. Miller, "The masking of speech," *Psychological Bulletin*, vol. 44, pp. 105-129, 1947.

[124] G.A. Miller and G.A. Heise, "The trill threshold," *J. Acoust. Soc. Am.*, vol. 22, pp. 637-638, 1950.

[125] A.R. Moghimi and R.M. Stern, "An analysis of binaural spectro-temporal masking as nonlinear beamforming," in *Proceedings of ICASSP*, pp. 835-839, 2014.

[126] B.C.J. Moore, *An introduction to the psychology of hearing.* 5th ed., San Diego CA: Academic Press, 2003.

[127] B.C.J. Moore, *Cochlear hearing loss.* 2nd ed., Chichester UK: Wiley, 2007.

[128] V. Nair and G.E. Hinton, "Rectified linear units improve restricted Boltzmann machines," in *Proceedings of ICML*, pp. 807–814, 2010.

[129] T. Nakatani, M. Ito, T. Higuchi, S. Araki, and K. Kinoshita, "Integrating DNN-based and spatial clustering-based mask estimation for robust MVDR beamforming," in *Proceedings of ICASSP*, pp. 286-290, 2017.

[130] A. Narayanan and D.L. Wang, "Ideal ratio mask estimation using deep neural networks for robust speech recognition," in *Proceedings of ICASSP*, pp. 7092-7096, 2013.

[131] P.A. Naylor and N.D. Gaubitch, Ed., *Speech dereverberation*. London: Springer, 2010.

[132] S. Nie, H. Zhang, X. Zhang, and W. Liu, "Deep stacking networks with time series for speech separation," in *Proceedings of ICASSP*, pp. 6717-6721, 2014.

[133] A.A. Nugraha, A. Liutkus, and E. Vincent, "Multichannel audio source separation with deep neural networks," *IEEE/ACM Trans. Audio Speech Lang. Proc.*, vol. 24, pp. 1652-1664, 2016.

[134] P. Papadopoulos, A. Tsiartas, and S. Narayanan, "Long-term SNR estimation of speech signals in known and unknown channel conditions," *IEEE/ACM Trans. Audio Speech Lang. Proc.*, vol. 24, pp. 2495-2506, 2016.

[135] A. Parbery-Clark, E. Skoe, C. Lam, and N. Kraus, "Musician enhancement for speech-in-noise," *Ear Hear.*, vol. 30, pp. 653-661, 2009.

[136] S.R. Park and J.W. Lee, "A fully convolutional neural network for speech enhancement," *arXiv:1609.07132v1*, 2016.

[137] R. Pascanu, T. Mikolov, and Y. Bengio, "On the difficulty of training recurrent neural networks," in *Proceedings of ICML*, pp. 1310–1318, 2013.

[138] S. Pascual, A. Bonafonte, and J. Serra, "SEGAN: Speech enhancement generative adversarial network," in *Proceedings of Interspeech*. pp. 3642-3646, 2017.

[139] L. Pfeifenberger, Zohrer, and F. Pernkopf, "DNN-based speech mask estimation for eigenvector beamforming," in *Proceedings of ICASSP*, pp. 66-70, 2017.

[140] A. Rix, J. Beerends, M. Hollier, and A. Hekstra, "Perceptual evaluation of speech quality (PESQ) - a new method for speech quality assessment of telephone networks and codecs," in *Proceedings of ICASSP*, pp. 749-752, 2001.

[141] N. Roman, D.L. Wang, and G.J. Brown, "Speech segregation based on sound localization," *J. Acoust. Soc. Am.*, vol. 114, pp. 2236-2252, 2003.

[142] F. Rosenblatt, *Principles of neural dynamics.* New York: Spartan, 1962.

[143] D.E. Rumelhart, G.E. Hinton, and R.J. Williams, "Learning internal representations by error propagation," in *Parallel distributed processing,* D.E. Rumelhart and J.L. McClelland, Ed., Cambridge, MA: MIT Press, pp. 318-362, 1986.

[144] S. Russell and P. Norvig, *Artificial intelligence: A modern approach.* 3rd ed., Upper Saddle River, NJ: Prentice Hall, 2010.

[145] M.R. Schadler, B.T. Meyer, and B. Kollmeier, "Spectro-temporal modulation subspace-spanning filter bank features for robust automatic speech recognition," *J. Acoust. Soc. Am.*, vol. 131, pp. 4134-4151, 2012.

[146] J. Schmidhuber, "Deep learning in neural networks: An overview," *Neural Netw.*, vol. 61, pp. 85-117, 2015.

[147] M.L. Seltzer, B. Raj, and R.M. Stern, "A Bayesian classifier for spectrographic mask estimation for missing feature speech recognition," *Speech Comm.*, vol. 43, pp. 379-393, 2004.

[148] S. Shamma, M. Elhilali, and C. Micheyl, "Temporal coherence and attention in auditory scene analysis," *Trends in Neuroscience*, vol. 34, pp. 114-123, 2011.

[149] B.J. Shannon and K.K. Paliwal, "Feature extraction from higher-order autocorrelation coefficients for robust speech recognition," *Speech Comm.*, vol. 48, pp. 1458-1485, 2006.

[150] Y. Shao, S. Srinivasan, and D.L. Wang, "Robust speaker identification using auditory features and computational auditory scene analysis," in *Proceedings of ICASSP*, pp. 1589-1592, 2008.

[151] B. Shinn-Cunningham, "Object-based auditory and visual attention," *Trends in Cognitive Sciences*, vol. 12, pp. 182-186, 2008.

[152] S. Srinivasan, N. Roman, and D.L. Wang, "Binary and ratio time-frequency masks for robust speech recognition," *Speech Comm.*, vol. 48, pp. 1486-1501, 2006.

[153] R.K. Srivastava, K. Greff, and J. Schmidhuber, "Highway networks," *arXiv1505.00387*, 2015.

[154] V. Summers and M.R. Leek, "F0 processing and the separation of competing speech signals by listeners with normal hearing and with hearing loss," *Journal of Speech, Language, and Hearing Research*, vol. 41, pp. 1294-1306, 1998.

[155] L. Sun, J. Du, L.-R. Dai, and C.-H. Lee, "Multiple-target deep learning for LSTM-RNN based speech enhancement," in *Proceedings of the Workshop on Hands-free Speech Communication and Microphone Arrays*, pp. 136-140, 2017.

[156] M. Sundermeyer, H. Ney, and R. Schluter, "From feedforward to recurrent LSTM neural networks for language modeling," *IEEE/ACM Trans. Audio Speech Lang. Proc.*, vol. 23, pp. 517–529, 2015.

[157] I. Sutskever, O. Vinyals, and Q.V. Le, "Sequence to sequence learning with neural networks," in *Proceedings of NIPS*, pp. 3104–3112, 2014.

[158] C.H. Taal, R.C. Hendriks, R. Heusdens, and J. Jensen, "An algorithm for intelligibility prediction of time-frequency weighted noisy speech," *IEEE Trans. Audio Speech Lang. Proc.*, vol. 19, pp. 2125-2136, 2011.

[159] D. Tabri, K.M. Chacra, and T. Pring, "Speech perception in noise by monolingual, bilingual and trilingual listeners," *International Journal of Language and Communication Disorders*, vol. 46, pp. 411-422, 2011.

[160] S. Tamura and A. Waibel, "Noise reduction using connectionist models," in *Proceedings of ICASSP*, pp. 553-556, 1988.

[161] M. Tu and X. Zhang, "Speech enhancement based on deep neural networks with skip connections," in *Proceedings of ICASSP*, pp. 5565-5569, 2017.

[162] Y. Tu, J. Du, Y. Xu, L.-R. Dai, and C.-H. Lee, "Speech separation based on improved deep neural networks with dual

outputs of speech features for both target and interfering speakers," in *Proceedings of ISCSLP*, pp. 250-254, 2014.
[163] L.P.A.S. van Noorden, *Temporal coherence in the perception of tone sequences*. Ph.D. Dissertation, Eindhoven University of Technology, 1975.
[164] B.D. van Veen and K.M. Buckley, "Beamforming: A versatile approach to spatial filtering," *IEEE ASSP Magazine*, 4-24, April 1988.
[165] E. Vincent, R. Gribonval, and C. Fevotte, "Performance measurement in blind audio source separation," *IEEE Trans. Audio Speech Lang. Proc.*, vol. 14, pp. 1462-1469, 2006.
[166] T. Virtanen, J.F. Gemmeke, and B. Raj, "Active-set Newton algorithm for overcomplete non-negative representations of audio," *IEEE/ACM Trans. Audio Speech Lang. Proc.*, vol. 21, pp. 2277-2289, 2013.
[167] T.T. Vu, B. Bigot, and E.S. Chng, "Combining non-negative matrix factorization and deep neural networks for speech enhancement and automatic speech recognition," in *Proceedings of ICASSP*, pp. 499-503, 2016.
[168] D.L. Wang, "On ideal binary mask as the computational goal of auditory scene analysis," in *Speech separation by humans and machines,* P. Divenyi, Ed., Norwell MA: Kluwer Academic, pp. 181-197, 2005.
[169] D.L. Wang, "The time dimension for scene analysis," *IEEE Trans. Neural Net.*, vol. 16, pp. 1401-1426, 2005.
[170] D.L. Wang, "Time-frequency masking for speech separation and its potential for hearing aid design," *Trend. Amplif.*, vol. 12, pp. 332-353, 2008.
[171] D.L. Wang, "Deep learning reinvents the hearing aid," *IEEE Spectrum*, 32-37, March 2017.
[172] D.L. Wang and G.J. Brown, Ed., *Computational auditory scene analysis: Principles, algorithms, and applications*. Hoboken NJ: Wiley & IEEE Press, 2006.
[173] D.L. Wang, U. Kjems, M.S. Pedersen, J.B. Boldt, and T. Lunner, "Speech intelligibility in background noise with ideal binary time-frequency masking," *J. Acoust. Soc. Am.*, vol. 125, pp. 2336-2347, 2009.
[174] Y. Wang, J. Du, L.-R. Dai, and C.-H. Lee, "A gender mixture detection approach to unsupervised single-channel speech separation based on deep neural networks," *IEEE/ACM Trans. Audio Speech Lang. Proc.*, vol. 25, pp. 1535-1546, 2017.
[175] Y. Wang, J. Du, L.-R. Dai, and C.-H. Lee, "A maximum likelihood approach to deep neural network based nonlinear spectral mapping for single-channel speech separation," in *Proceedings of Interspeech*, pp. 1178-1182, 2017.
[176] Y. Wang, P. Getreuer, T. Hughes, R.F. Lyon, and R.A. Saurous, "Trainable frontend for robust and far-field keyword spotting," in *Proceedings of ICASSP*, pp. 5670-5674, 2017.
[177] Y. Wang, K. Han, and D.L. Wang, "Exploring monaural features for classification-based speech segregation," *IEEE Trans. Audio Speech Lang. Proc.*, vol. 21, pp. 270-279, 2013.
[178] Y. Wang, A. Narayanan, and D.L. Wang, "On training targets for supervised speech separation," *IEEE/ACM Trans. Audio Speech Lang. Proc.*, vol. 22, pp. 1849-1858, 2014.
[179] Y. Wang and D.L. Wang, "Boosting classification based speech separation using temporal dynamics," in *Proceedings of Interspeech*, pp. 1528-1531, 2012.
[180] Y. Wang and D.L. Wang, "Cocktail party processing via structured prediction," in *Proceedings of NIPS*, pp. 224-232, 2012.
[181] Y. Wang and D.L. Wang, "Towards scaling up classification-based speech separation," *IEEE Trans. Audio Speech Lang. Proc.*, vol. 21, pp. 1381-1390, 2013.
[182] Y. Wang and D.L. Wang, "A deep neural network for time-domain signal reconstruction," in *Proceedings of ICASSP*, pp. 4390-4394, 2015.
[183] Z.-Q. Wang and D.L. Wang, "Phoneme-specific speech separation," in *Proceedings of ICASSP*, pp. 146-150, 2016.
[184] Z. Wang, X. Wang, X. Li, Q. Fu, and Y. Yan, "Oracle performance investigation of the ideal masks," in *Proceedings of IWAENC*, 2016.
[185] F. Weninger*, et al.*, "Speech enhancement with LSTM recurrent neural networks and its application to noise-robust ASR," in *Proceedings of LVA/ICA*, pp. 91–99, 2015.
[186] F. Weninger, J. Hershey, J. Le Roux, and B. Schuller, "Discriminatively trained recurrent neural networks for single-channel speech separation," in *Proceedings of GlobalSIP*, pp. 740-744, 2014.
[187] P.J. Werbos, "Backpropagation through time: What it does and how to do it," *Proc. IEEE*, vol. 78, pp. 1550-1560, 1990.
[188] D.S. Williamson, Y. Wang, and D.L. Wang, "Complex ratio masking for monaural speech separation," *IEEE/ACM Trans. Audio Speech Lang. Proc.*, vol. 24, pp. 483–492, 2016.
[189] D.H. Wolpert, "The lack of *a priori* distinction between learning algorithms," *Neural Comp.*, vol. 8, pp. 1341-1390, 1996.
[190] B. Wu, K. Li, M. Yang, and C.-H. Lee, "A reverberation-time-aware approach to speech dereverberation based on deep neural networks," *IEEE/ACM Trans. Audio Speech Lang. Proc.*, vol. 25, pp. 102-111, 2017.
[191] M. Wu and D.L. Wang, "A two-stage algorithm for one-microphone reverberant speech enhancement," *IEEE Trans. Audio Speech Lang. Proc.*, vol. 14, pp. 774-784, 2006.
[192] S. Xia, H. Li, and X. Zhang, "Using optimal ratio mask as training target for supervised speech separation," in *Proceedings of APSIPA*, 2017.
[193] X. Xiao, S. Zhao, D.L. Jones, E.S. Chng, and H. Li, "On time-frequency mask estimation for MVDR beamforming with application in robust speech recognition," in *Proceedings of ICASSP*, pp. 3246-3250, 2017.
[194] X. Xiao*, et al.*, "Speech dereverberation for enhancement and recognition using dynamic features constrained deep neural networks and feature adaptation," *EURASIP J. Adv. Sig. Proc.*, vol. 2016, pp. 1-18, 2016.
[195] Y. Xu, J. Du, L.-R. Dai, and C.-H. Lee, "Dynamic noise aware training for speech enhancement based on deep neural networks," in *Proceedings of Interspeech*, pp. 2670-2674, 2014.
[196] Y. Xu, J. Du, L.-R. Dai, and C.-H. Lee, "An experimental study on speech enhancement based on deep neural networks," *IEEE Sig. Proc. Lett.*, vol. 21, pp. 65-68, 2014.
[197] Y. Xu, J. Du, L.-R. Dai, and C.-H. Lee, "A regression approach to speech enhancement based on deep neural networks," *IEEE/ACM Trans. Audio Speech Lang. Proc.*, vol. 23, pp. 7-19, 2015.
[198] Y. Xu, J. Du, Z. Huang, L.-R. Dai, and C.-H. Lee, "Multi-objective learning and mask-based post-processing for deep neural network based speech enhancement," in *Proceedings of Interspeech*, pp. 1508-1512, 2015.
[199] O. Yilmaz and S. Rickard, "Blind separation of speech mixtures via time-frequency masking," *IEEE Trans. Sig. Proc.*, vol. 52, pp. 1830-1847, 2004.
[200] T. Yoshioka, M. Ito, M. Delcroix, and e. al., "The NTT CHiME-3 system: advances in speech enhancement and recognition for mobile multi-microphone devices," in *Proceedings of IEEE ASRU*, 2015.
[201] W.A. Yost, "The cocktail party problem: Forty years later," in *Binaural and spatial hearing in real and virtual environments,* R.H. Gilkey and T.R. Anderson, Ed., Mahwah, NJ: Lawrence Erlbaum, pp. 329-347, 1997.
[202] D. Yu, M. Kolbak, Z.-H. Tan, and J. Jensen, "Permutation invariant training of deep models for speaker-independent

multi-talker speech separation," in *Proceedings of ICASSP*, pp. 241-245, 2017.

[203] Y. Yu, W. Wang, and P. Han, "Localization based stereo speech source separation using probabilistic time-frequency masking and deep neural networks," *EURASIP J. Audio Speech Music Proc.*, vol. 2016, pp. 1-18, 2016.

[204] K.-H. Yuo and H.-C. Wang, "Robust features for noisy speech recognition based on temporal trajectory filtering of short-time autocorrelation sequences," *Speech Comm.*, vol. 28, pp. 13-24, 1999.

[205] H. Zhang, X. Zhang, and G. Gao, "Multi-target ensemble learning for monaural speech separation," in *Proceedings of Interspeech*, pp. 1958-1962, 2017.

[206] X.-L. Zhang and D.L. Wang, "A deep ensemble learning method for monaural speech separation," *IEEE/ACM Trans. Audio Speech Lang. Proc.*, vol. 24, pp. 967-977, 2016.

[207] X.-L. Zhang and J. Wu, "Deep belief networks based voice activity detection," *IEEE Trans. Audio Speech Lang. Proc.*, vol. 21, pp. 697-710, 2013.

[208] X. Zhang and D.L. Wang, "Deep learning based binaural speech separation in reverberant environments," *IEEE/ACM Trans. Audio Speech Lang. Proc.*, vol. 25, pp. 1075-1084, 2017.

[209] X. Zhang, Z.-Q. Wang, and D.L. Wang, "A speech enhancement algorithm by iterating single- and multi-microphone processing and its application to robust ASR," in *Proceedings of ICASSP*, pp. 276-280, 2017.

[210] X. Zhang, H. Zhang, S. Nie, G. Gao, and W. Liu, "A pairwise algorithm using the deep stacking network for speech separation and pitch estimation," *IEEE/ACM Trans. Audio Speech Lang. Proc.*, vol. 24, pp. 1066-1078, 2016.

[211] Y. Zhao, Z.-Q. Wang, and D.L. Wang, "A two-stage algorithm for noisy and reverberant speech enhancement," in *Proceedings of ICASSP*, pp. 5580-5584, 2017.